\pdfoutput=1

\documentclass[a4paper,twoside]{article}

\usepackage{epsfig}
\usepackage{subcaption}
\usepackage{calc}
\usepackage{amssymb}
\usepackage{amstext}
\usepackage{amsmath}
\usepackage{amsthm}
\usepackage{multicol}
\usepackage{pslatex}
\usepackage{apalike}
\usepackage{algorithm2e}
\usepackage[bottom]{footmisc}

\usepackage[export]{adjustbox}
\usepackage{graphicx}
\graphicspath{{./fig/}}
\DeclareGraphicsExtensions{.pdf,.jpeg,.png,.jpg}
\usepackage{epsfig}
\usepackage{amsmath}
\usepackage{amssymb}
\usepackage{booktabs}
\usepackage{colortbl}
\usepackage{textcomp}
\usepackage{subcaption}
\usepackage{standalone}
\usepackage[dvipsnames,x11names]{xcolor}
\usepackage{pgfplots} 
\usepackage{multirow}
\usepackage[inkscapelatex=false]{svg}
\usepackage{siunitx}
\usepackage{lipsum}
\usepackage{pifont}
\usepackage{yfonts}
\usepackage{listings}
\usepackage{makecell}
\usepackage[T1]{fontenc}
\usepackage[normalem]{ulem}
\usepackage{nicematrix}
\usepackage{placeins}
\usepackage{circledsteps}
\usepackage{scalerel}
\usepackage{todonotes}
\usepackage{ulem}
\usepackage{appendix}
\usepackage{comment}
\usepackage{amsthm}
\usepackage{amsbsy}
\usepackage{booktabs}
\usepackage{amssymb}
\usepackage{graphicx}
\usepackage{multirow}
\usepackage{svg}
\usepackage{siunitx}
\usepackage{pgfplots}
\usepackage{amsmath}
\usepackage{hyperref}
\usepackage{makecell}
\usepackage{pifont}
\usepackage{hhline}
\usepackage{mathtools}
\usepackage{bbold}
\usepackage{xurl}
\newcommand{\cmark}{\ding{51}}%
\newcommand{\xmark}{\ding{55}}%
\usepackage{tikz}
\usepackage[capitalize]{cleveref}
\crefname{section}{Sec.}{Secs.}
\Crefname{section}{Section}{Sections}
\Crefname{table}{Table}{Tables}
\crefname{table}{Tab.}{Tabs.}
\definecolor{ForestGreen}{RGB}{34,170,34}
\definecolor{textcolortab}{RGB}{70,170,34}
\definecolor{ao(english)}{rgb}{0.0, 0.5, 0.0}
\pgfplotsset{compat=1.18}

\usepackage{SCITEPRESS}     

\begin{document}

\title{MetaToken: Detecting Hallucination \\
in Image Descriptions by Meta Classification}

\author{\authorname{Laura Fieback\sup{1,2}, Jakob Spiegelberg\sup{1} and Hanno Gottschalk\sup{2}}
\affiliation{\sup{1}Volkswagen AG, Berliner Ring 2, 38440 Wolfsburg, Germany}
\affiliation{\sup{2}Mathematical Modeling of Industrial Life Cycles, Institute of Mathematics, TU Berlin, Berlin, Germany}
\email{\{laura.fieback, jakob.spiegelberg\}@volkswagen.de, gottschalk@math.tu-berlin.de}
}


\abstract{Large Vision Language Models (LVLMs) have shown remarkable capabilities in multimodal tasks like visual question answering or image captioning. However, inconsistencies between the visual information and the generated text, a phenomenon referred to as hallucinations, remain an unsolved problem with regard to the trustworthiness of LVLMs. To address this problem, recent works proposed to incorporate computationally costly Large (Vision) Language Models in order to detect hallucinations on a sentence- or subsentence-level. In this work, we introduce \textbf{MetaToken}, a lightweight binary classifier to detect hallucinations on the token-level at negligible cost. Based on a statistical analysis, we reveal key factors of hallucinations in LVLMs. MetaToken can be applied to any open-source LVLM without any knowledge about ground truth data providing a calibrated detection of hallucinations. We evaluate our method on four state-of-the-art LVLMs demonstrating the effectiveness of our approach.}

\onecolumn \maketitle \normalsize \setcounter{footnote}{0} \vfill

\section{\uppercase{Introduction}}
\label{sec:intro}
LVLMs have demonstrated an impressive visual-language understanding by aligning text and visual features. However, besides their remarkable zero-shot performance on visual downstream tasks, LVLMs suffer from the problem of hallucinations \cite{Li.2023,Liu.2024,Rohrbach.2018} inherited from the underlying LLMs or caused by faulty interpretation of the image input by the vision branch. In the context of LVLMs, hallucination refers to the problem of inconsistencies between the generated text and the visual input \cite{Liu.2024}, diminishing the trustworthiness of these models. Especially in safety-critical applications like autonomous driving \cite{Gao.2024,Tian.2024} or medicine \cite{Jiang.2024,Li.2023c}, the reliability of the underlying model is indispensable for decision making. In order to address this problem, recent works \cite{liu2024mitigating,Gunjal.2023,Zhao.2024,Dai.2023,Xing.2024} have proposed additional instruction tuning datasets and pre-training strategies to detect and mitigate hallucinations on a sentence- or subsentence-level. Another common strategy comprises stacked L(V)LMs to post-hoc detect and rectify hallucinations \cite{Wu.2024,Yin.2023,Jing.2023}.

\begin{figure}[t!]
  \centering
  \includegraphics[width=\columnwidth]{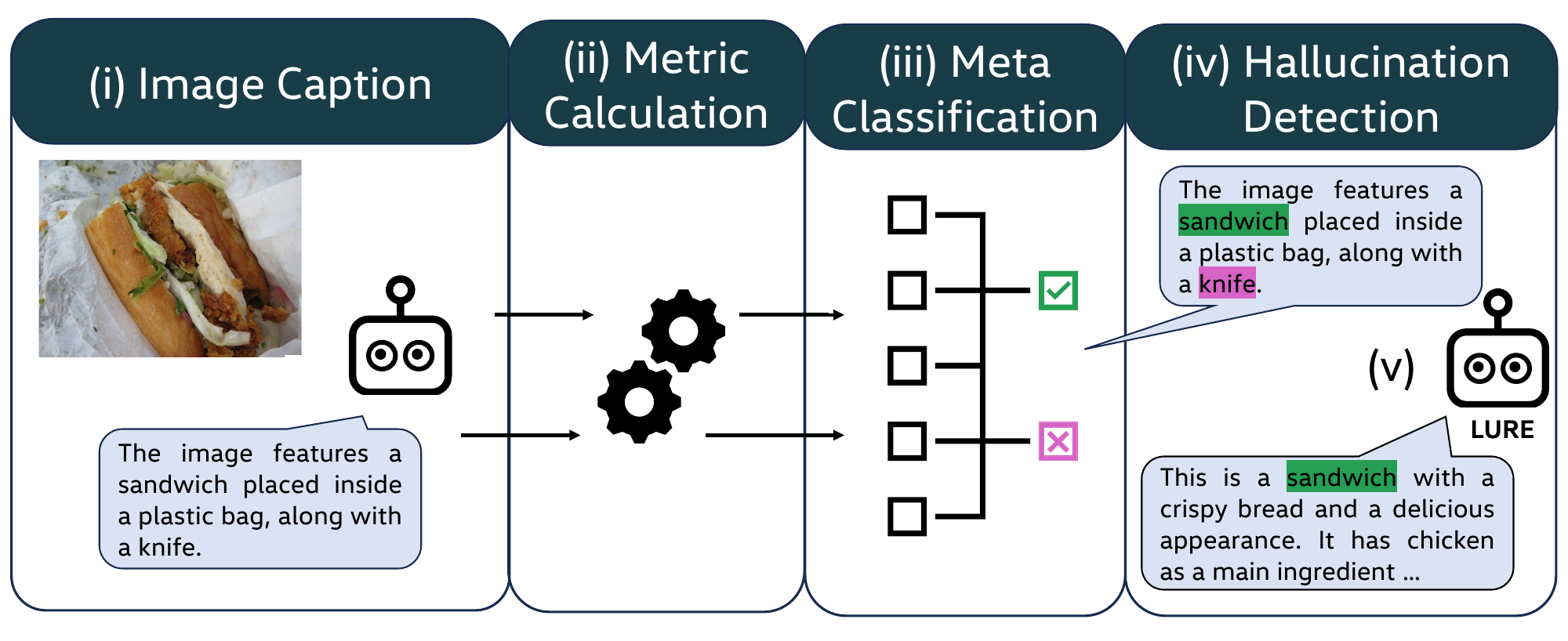}
  \caption{\textbf{MetaToken.} Based on generated image captions (i), we calculate our proposed input features (ii) (see \cref{subsec:metrics}). Afterwards, we apply the trained meta classifier (iii) to detect \colorbox{Mulberry!50}{hallucinated} and \colorbox{ao(english)!50}{true} objects (iv). Moreover, (v) MetaToken can be easily integrated into LURE \cite{Zhou.2023} to improve the hallucination mitigation.}
  \label{fig:metatoken}
\end{figure}

\begin{table*}[h]
  \centering
  \caption{\textbf{Related Work on Hallucination Detection.} A comparison of existing approaches on hallucination detection with respect to computational efficiency, i.e., whether the respective method can be implemented without an additional dataset, fine-tuning or prompting an L(V)LM. \textcolor{green}{\cmark} indicates 'yes', \textcolor{red}{\xmark} indicates 'no'.}
  
   \begin{tabular}{cccc}
        \toprule
        method & w/o add. dataset & w/o fine-tuning & w/o prompting \\
        \midrule
        LogicCheckGPT \cite{Wu.2024} & \textcolor{green}{\cmark} & \textcolor{green}{\cmark} & \textcolor{red}{\xmark} \\
        Woodpecker \cite{Yin.2023} &  \textcolor{green}{\cmark} & \textcolor{green}{\cmark} & \textcolor{red}{\xmark} \\
        M-HalDetect \cite{Gunjal.2023} & \textcolor{red}{\xmark} & \textcolor{red}{\xmark} & \textcolor{red}{\xmark} \\
        HaELM \cite{Wang.2023} &  \textcolor{red}{\xmark} & \textcolor{red}{\xmark} & \textcolor{red}{\xmark} \\
        FAITHSCORE \cite{Jing.2023} & \textcolor{green}{\cmark} & \textcolor{green}{\cmark} & \textcolor{red}{\xmark} \\
        UNIHD \cite{Chen.2024} & \textcolor{green}{\cmark} & \textcolor{green}{\cmark} & \textcolor{red}{\xmark} \\
        Ours & \textcolor{green}{\cmark} & \textcolor{green}{\cmark} & \textcolor{green}{\cmark} \\
        \bottomrule  
    \end{tabular}

  \label{tab:related_work} 
\end{table*}

In this work, we tackle the problem of object hallucination in image captions. To this end, we introduce MetaToken, a lightweight hallucination detection method which can be applied to any open-source LVLM. MetaToken builds up on the idea of meta classification \cite{Lin.2003,hendrycks17baseline,Chen.2019,Rottmann.2020,Fieback.2023} to detect hallucinated objects on the token-level based on the model output only. \cref{fig:metatoken} depicts our approach. In contrast to existing methods, our approach neither requires an additional dataset, fine-tuning an L(V)LM nor cost-intensive L(V)LM prompting. Within a comprehensive statistical analysis, we investigate a broad set of input features which are indicative of hallucinations providing deep insights into the sources of this specific type of model errors. We evaluate our method on four state-of-the-art (SOTA) LVLMs \cite{Dai.2023b,Ye.2023,Zhu.2023,Huang.2023} achieving area under receiver operator characteristic curve values \cite{Davis.2006} of up to $92.12 \%$ and area under precision recall curve values \cite{Davis.2006} of up to $84.01 \%$. Moreover, we show that our method can be incorporated into the LURE mitigation method \cite{Zhou.2023}. While LURE reduces hallucinations by up to $52.98 \%$, we achieve a hallucination reduction by up to $56.62 \%$ through the superior precision-recall-ratio of MetaToken. Our main contributions are as follows:
\begin{itemize}
    \item We propose and investigate a comprehensive set of statistics as potential factors of object hallucinations.
    \item Based on these statistics, we introduce MetaToken, a lightweight binary classifier to detect object hallucinations as a post-hoc method. MetaToken can be applied to any open-source LVLM without any knowledge about the ground truth data.
    \item We show that MetaToken can be easily integrated into the LURE mitigation method, outperforming the initial LURE results through a superior precision-recall-ratio.
\end{itemize}

\section{\uppercase{Related Work}}
\label{sec:related_work}
\subsection{Hallucinations in LVLMs}
Hallucinations in LVLMs can occur on different semantic levels, where coarse-grained object hallucination \cite{Rohrbach.2018} refers to objects generated in the language output, which are not depicted in the input image, whereas fine-grained hallucination describes inconsistencies with respect to object attributes or relations between objects \cite{Li.2023,Liu.2024}. For a comprehensive survey on hallucinations in LVLMs, we refer to \cite{Liu.2024}.

The problem of hallucination mitigation is mainly tackled by either retraining the model with an instruction tuning dataset \cite{liu2024mitigating,Gunjal.2023}, rectifying image captions as a post-processing step or incorporating new pre-training or generation strategies. LURE \cite{Zhou.2023} serves as a post-hoc method to rectify object hallucinations by training an LVLM-based revisor to reconstruct less hallucinatory descriptions. MARINE \cite{Zhao.2024} enriches the visual context of LVLMs by incorporating object grounding features into the LLM input. In \cite{Dai.2023}, a new pre-training objective is introduced to mitigate object hallucinations by improving the object-level image-text alignment. Since the reliability of the generated language output still remains an unsolved problem, many researches focus on the problem of hallucination detection. In \cite{Wu.2024}, the problem of hallucination detection and mitigation is solved simultaneously by raising logical correlated questions and checking for logical consistency throughout the generated answers afterwards. Similarly, Woodpecker \cite{Yin.2023} serves as a post-processing hallucination detection and correction method incorporating visual knowledge validation for both instance- and attribute-level hallucinations. A human-labeled dataset is published in \cite{Gunjal.2023}, which is used to train a binary classifier based on the LVLM InstructBLIP \cite{Dai.2023b} to classify between accurate and inaccurate sentences. In \cite{Wang.2023}, an LLM-based hallucination evaluation framework is introduced by training an LLM to distinguish between hallucinated and hallucination-free image captions. Both, \cite{Jing.2023} and \cite{Chen.2024}, propose a pipeline consisting of several LVLMs and LLMs to verify each claim contained in the generated language output. In this work, we tackle the problem of object hallucination detection using meta classification, that is, training a lightweight binary classifier to distinguish between true and hallucinated objects. In contrast to existing methods, our approach neither requires an additional dataset, fine-tuning an L(V)LM nor cost-intensive L(V)LM prompting for claim verification (see \cref{tab:related_work}).

\subsection{Meta Classification}
In classical machine learning, meta classification refers to the problem of how to best combine predictions from an ensemble of classifiers \cite{Lin.2003}. In terms of deep learning, this concept has been transferred to the classification whether a prediction is true or false based on uncertainty features \cite{hendrycks17baseline}. Several works have applied this idea to natural language processing \cite{Vasudevan.2019,Liu.2022,Gui.2024}, image classification \cite{Chen.2019}, semantic segmentation \cite{Rottmann.2020,Rottmann.2019,Maag.2020,Fieback.2023}, video instance segmentation \cite{Maag.2021} and object detection \cite{Schubert.2021,Kowol.2020}. We are the first to transfer the idea of meta classification to the problem of hallucination detection for LVLMs. Based on a statistical analysis of key factors of hallucinations in LVLMs, we identify input features outperforming classical uncertainty-based statistics.

\subsection{Hallucination Evaluation}
Since different studies \cite{Rohrbach.2018,Dai.2023} have shown that standard image captioning metrics like BLEU \cite{Papineni.2002}, METEOR \cite{Lavie.2007}, CIDEr \cite{Vedantam.2015} and SPICE \cite{Anderson.2016} are not capable of measuring hallucinations properly, most works on hallucination mitigation measure the performance of their proposed method in terms of the Caption Hallucination Assessment with Image Relevance (CHAIR) metric \cite{Rohrbach.2018}. The CHAIR metric measures the proportion of hallucinated objects in an image caption by matching the objects in the generated text against the ground truth annotations \cite{Lin.2014}. Thus, for every generated object, the CHAIR metric provides a binary label indicating whether the object is true, i.e., contained in the image, or hallucinated. Further datasets and evaluation methods have been proposed to evaluate the performance of LVLMs across multiple multi-modal tasks \cite{Fu.2023,Gunjal.2023,Lovenia.2023,Liu.2023b}. While some of the proposed evaluation methods ask LLMs to output quality-related scores \cite{Liu.2023,liu2024mitigating,Yu.2023} or measure the image-text similarity \cite{Hessel.2021}, other methods \cite{Li.2023,Fu.2023,Wang.2024} use a prompt template to query hallucination-related questions and force the model to answer either \textit{'yes'} or \textit{'no'}. However, the results in \cite{Li.2023} and \cite{Fu.2023} have shown that LVLMs tend to answer \textit{'yes'}, which results in a low recall for hallucinated objects. Moreover, the LLM- and similarity-based scores \cite{Liu.2023,liu2024mitigating,Yu.2023,Hessel.2021} evaluate the entire image caption in terms of continuous scores instead of providing a binary label for each generated object. Thus, we rely on the CHAIR metric to evaluate hallucinations.
\section{\uppercase{Method}}
\label{sec:method}
The aim of our method is to detect hallucinations in the text output of LVLMs leveraging the idea of meta classification. To this end, we build input features based on the model output that have been shown to correlate with hallucinations. These features are used to train a lightweight binary meta model to classify between hallucinated and true objects. At inference time, we can detect hallucinations by computing the proposed features and applying the trained meta model afterwards. A formal definition of meta classification is provided in \cref{subsec:haldetectMC}.

\subsection{Notation}
Typically, LVLMs generate language output in an auto-regressive manner by predicting the probability distribution of the next token over the entire vocabulary $\mathcal{V}$ given the input image $x$, the provided prompt $q$ as well as the already generated tokens. For this purpose, the image $x$ as well as the prompt $q$ are tokenized into $u+1$ image tokens $t_{x_0}, \dots, t_{x_u}$ and $v+1$ prompt tokens $t_{q_0}, \dots, t_{q_v}$, respectively. 

We denote the sequence of generated output tokens by $s=(t_0,\dots,t_K)$ with sequence length $K+1$. Moreover, let $s_i=(t_0,\dots,t_i)$ denote the generated output sequence at generation step $i$. The probability of generating token $t_{i+1} \in \mathcal{V}$ at generation step $i+1$ given the input image $x$, the provided prompt $q$ and the already generated tokens $s_i$ can be formulated as $p(t_{i+1}|x,q,s_i)$. For a shorter notation, we define $\hat{p}_{i+1} = p(t_{i+1}|x,q,s_i)$. Furthermore, let $p_{i+1}$ denote the probability distribution at generation step $i+1$ over the dictionary $\mathcal{V}$ and $|\mathcal{V}|$ the cardinality of $\mathcal{V}$.

Given the language output $s$, we extract all objects contained in the generated text. We denote the set of objects contained in the sequence $s$ by $\mathcal{O}_s=\{o_0,\dots,o_z \}$. Since the generated string of an object might consist of several tokens, we define for every object $o_j \in \mathcal{O}_s$ the start token $t_{o_{j,s}}$ at position $0 \leq o_{j,s} \leq K$ as well as the end token $t_{o_{j,e}}$ at position $0 \leq o_{j,e} \leq K$.

\subsection{Input Features} \label{subsec:metrics}
Recent works \cite{Rohrbach.2018,Wang.2023} have investigated influencing factors of object hallucinations. First, the results in \cite{Wang.2023} indicate that LVLMs often generate true segments at the beginning while the risk of hallucinations increases at the letter part of the generated responses. Thus, we take account of the relative position (\cref{eq:relpos}) of a generated object and the absolute occurrence (\cref{eq:absocc}) of the object in the generated text. Second, to account for the over-reliance of LVLMs on language priors during the generation process \cite{Rohrbach.2018,Wang.2023}, we consider the mean absolute attention on the image tokens (\cref{eq:att}). Finally, we regard the model uncertainty through different dispersion measures (\cref{eq:logp}-(\ref{eq:probdiff})) which have been shown to correlate with model errors in different fields \cite{Rottmann.2019,Schubert.2021,Vasudevan.2019} including the sequence score\footnote{\url{https://huggingface.co/docs/transformers/main/en/main_classes/text_generation\#transformers.GenerationMixin.compute_transition_scores}} calculated during the LVLM generation process. For a generated object $o_j \in \mathcal{O}_s$ from the output sequence $s=(t_0,\dots,t_K)$, we define

\begin{itemize}
    \item the \textbf{relative position}
    \begin{equation} \label{eq:relpos}
        P_{o_j} = o_{j,s} / (K+1),
    \end{equation}
    \item the \textbf{absolute occurrence} of object $o_j$ in $s$
    \begin{equation} \label{eq:absocc}
       N_{o_j} = \sum_{l=0} ^{z} \mathbb{1}_{\{ o_l = o_j \}}, 
    \end{equation}
    \item for every attention head $g = 0,\dots, G-1$, the \textbf{mean absolute attention} of the start token $t_{o_{j,s}}$ on the image tokens $t_{x_0}, \dots, t_{x_u}$
    \begin{equation} \label{eq:att}
        A_{o_j}^{g} = \frac{1}{u+1} \sum_{r=0} ^{u} |\textrm{Attention}_{t_{o_{j,s}}}({t_{x_r}})|,
    \end{equation}
    where $\textrm{Attention}_{t_{i}}({t_{n}})$ denotes the attention of the generated token $t_i$ at generation step $i$ on the input token $t_n$,
    \item the \textbf{log probability}
    \begin{equation} \label{eq:logp}
        L_{o_j} = \sum_{i=o_{j,s}} ^{o_{j,e}} \log \hat{p}_{i},
    \end{equation}
    \item the \textbf{cumulated log probability}
    \begin{equation} \label{eq:cumlogp}
        C_{o_j} = \sum_{i=0} ^{o_{j,e}} \log \hat{p}_{i},
    \end{equation}
    \item the sequence \textbf{score} with \textit{length\_penalty} parameter $l_p$
    \begin{equation} \label{eq:score}
        S_{o_j} = \frac{1}{(o_{j,e})^{l_p}} \sum_{i=0} ^{o_{j,e}} \log \hat{p}_{i},
    \end{equation}
    \item the \textbf{variance}
    \begin{equation}  \label{eq:var}
        V_{o_j} =  \frac{1}{|\mathcal{V}|} \sum_{t \in \mathcal{V}} (\log p_{o_{j,s}} (t) - \mu )^2
    \end{equation} 
    with \quad $\mu = \frac{1}{|\mathcal{V}|} \sum_{t \in \mathcal{V}} \log p_{o_{j,s}} (t)$,
    \item the \textbf{entropy} \cite{Shannon.1948}
    \begin{equation} \label{eq:entropy}
        E_{o_j} = - \frac{1}{\log |\mathcal{V}|} \sum_{t \in \mathcal{V}} p_{o_{j,s}} (t) \log p_{o_{j,s}} (t),
    \end{equation}
    \item the \textbf{variation ratio}
    \begin{equation} \label{eq:varrat}
        R_{o_j} = 1 - p_{o_{j,s}} (t_{\mathrm{max}}), \quad t_{\mathrm{max}} = \max_{t \in \mathcal{V}} p_{o_{j,s}} (t),
    \end{equation} 
    \item the \textbf{probability margin}
    \begin{equation} \label{eq:probmargin}
        M_{o_j} = R_{o_j} + \max_{t \in \mathcal{V}\setminus \{t_{\mathrm{max}}\}} p_{o_{j,s}} (t), \quad \textrm{and},
    \end{equation}
    \item the \textbf{probability difference}
    \begin{equation} \label{eq:probdiff}
        D_{o_j} = \log p_{o_{j,s}} (t_{\mathrm{max}}) - \log \hat{p}_{o_{j,s}}.
    \end{equation}
\end{itemize}

Finally, for an object $o_j \in \mathcal{O}_s$, we define the total set of input features as
\begin{equation} \label{eq:M}
\begin{split}
    \mathcal{M}_{o_j} = \{ & P_{o_j}, N_{o_j}, A^{0}_{o_j},\dots, A^{G-1}_{o_j}, L_{o_j}, \\
    & C_{o_j}, S_{o_j}, V_{o_j}, E_{o_j}, R_{o_j}, M_{o_j}, D_{o_j}\}
\end{split}
\end{equation}
with cardinality $|\mathcal{M}_{o_j}| = 10+G$.

\subsection{Hallucination Detection Using $\quad$ Meta Classification} \label{subsec:haldetectMC}
Let $\mathcal{M}$ denote a set of input features with cardinality $|\mathcal{M}|$. The idea of meta classification consists of training a lightweight binary meta model based on the input features $\mathcal{M}$ to classify between true and false predictions, i.e., to detect true and hallucinated objects in the generated output $s$. To this end, let
\begin{equation} \label{eq:metaclassifier}
    f:\mathbb{R}^{|\mathcal{M}|} \rightarrow \{0,1\}
\end{equation}
denote the binary classifier. We denote the set of training captions by $\mathcal{S}^{\mathrm{train}}$ and the corresponding set of generated objects by
\begin{equation}
    \mathcal{O}_{\mathcal{S}^{\mathrm{train}}} = \bigcup_{s \in \mathcal{S}^{\mathrm{train}}} \mathcal{O}_{s}.
\end{equation}
For every generated $o_j \in \mathcal{O}_{\mathcal{S}^{\mathrm{train}}}$ we build an input vector $m_{o_j} \in \mathbb{R}^{|\mathcal{M}_{o_j}|}$ representing the feature set $\mathcal{M}_{o_j}$ (\cref{eq:M}) and define the label $y_{o_j} \in \{0,1\}$ according to the CHAIR evaluation (see \cref{subseq:haleval}). After standardizing the inputs, we use the set
\begin{equation}
    \{(m_{o_j},y_{o_j}) \quad | \quad j=0,\dots, |\mathcal{O}_{\mathcal{S}^{train}}|-1\}
\end{equation}
to train the classifier $f$.

Given a generated caption $s$ at inference time, we calculate the input vector $m_{o_j} \in  \mathbb{R}^{|\mathcal{M}_{o_j}|}$ for every object $o_j \in \mathcal{O}_{s}$ and apply the trained binary meta classifier $f$ to detect hallucinated objects. Note that the input vector $m_{o_j}$ can be calculated in an automated manner based on the model output only without any knowledge of the ground truth data.

\begin{table}[t!]
  \centering
  \caption{\textbf{LVLM Performance.} Evaluation results of SOTA LVLMs with respect to the average number of generated objects per image ($\# \textrm{obj.}$) and $\textrm{CHAIR}_i$ based on nucleus sampling (s) \cite{Holtzman.2020} and beam search (b) \cite{Bisiani.1987}. The best results in each block are highlighted.}
  
   \resizebox{\columnwidth}{!}{
   \begin{tabular}{c|cc|cc}
        \toprule
        \multicolumn{1}{c}{} & \multicolumn{2}{c}{MSCOCO} & \multicolumn{2}{c}{BDD100K} \\
        Model & $\# \textrm{obj.}$ & $\textrm{CHAIR}_i \downarrow$ & $\# \textrm{obj.}$ & $\textrm{CHAIR}_i \downarrow$ \\
        \hline
        InstructBLIP (s) & $5.6$ & $\textbf{10.4}$ & $5.8$ & $\textbf{22.07}$ \\
        mPLUG-Owl (s) & $6.9$ & $30.2$ &$5.8$ & $33.44$ \\
        MiniGPT-4 (s) & $4.4$ & $13.6$ & $3.7$ & $29.12$ \\
        LLaVa (s) & $7.3$ & $18.7$ & $7.1$ & $29.72$ \\   
        \hline
        InstructBLIP (b) & $5.1$ & $\textbf{7.6}$ & - & - \\
        mPLUG-Owl (b) & $6.7$ & $25.3$ & - & - \\
    \end{tabular}
    }

  \label{tab:modeloverview}
\end{table}
\section{\uppercase{Experimental Settings}}
\label{sec:experiments}
\subsection{Datasets} \label{subseq:dataset}
We evaluate our method on the MSCOCO and BDD100K datasets. The MSCOCO dataset \cite{Lin.2014} is a large-scale dataset for object detection, segmentation, and image captioning comprising more than $200 K$ labeled images. The BDD100K dataset \cite{bdd100k} consists of $100 K$ labeled street scene images including labels for object detection, semantic segmentation and instance segmentation. We randomly sample $5,000$ images from the validation sets and produce image captions $s$ for four SOTA LVLMs. We use $80 \%$ of the generated captions as training set $\mathcal{O}_{\mathcal{S}^{\mathrm{train}}}$ and validate our method on the remaining $20 \%$ denoted as $\mathcal{O}_{\mathcal{S}^{\mathrm{val}}}$. In our experiments, we average our results over ten randomly sampled training-validation splits. The corresponding standard deviations are given in parentheses.

\subsection{Hallucination Evaluation} \label{subseq:haleval}

\begin{table}[t!]
  \centering
  \caption{\textbf{Expected Calibration Error.} The $ECE$ for the LR and GB MetaToken classifier.}
  
   \resizebox{.85\columnwidth}{!}{
   \begin{tabular}{c|c|c}
        \toprule
         \multicolumn{1}{c}{} & \multicolumn{2}{c}{$ECE \hspace{1.5mm} (\textrm{in} \hspace{1.5mm} \%) \hspace{1mm} \downarrow$} \\
        Model & LR & GB \\
        \hline
        InstructBLIP & $0.81 ^{(\pm 3.2\mathrm{e}-4)}$ & $1.29 ^{(\pm 4.6\mathrm{e}-4)}$ \\
        mPLUG-Owl & $1.36 ^{(\pm 1.3\mathrm{e}-3)}$ & $2.01 ^{(\pm 6.9\mathrm{e}-4)}$ \\
        MiniGPT-4 & $1.35 ^{(\pm 9.5\mathrm{e}-4)}$ & $1.43 ^{(\pm 8.8\mathrm{e}-4)}$ \\
        LLaVa & $1.05 ^{(\pm 2.0\mathrm{e}-4)}$ & $1.38 ^{(\pm 8.1\mathrm{e}-4)}$ \\
        \bottomrule    
    \end{tabular}
    }

   \label{tab:ece}
\end{table}

The CHAIR metric \cite{Rohrbach.2018} is an automated hallucination evaluation method which measures the proportion of hallucinated objects by matching the generated text against the ground truth objects. Given an image caption $s$, CHAIR provides a binary label for the generated objects $o_{j} \in \mathcal{O}_{s}$ indicating whether each object is true, i.e., contained in the image, or hallucinated. CHAIR is defined on two semantic levels: The object instance level $\textrm{CHAIR}_i$ to measure the proportion of hallucinated objects in an image caption which is defined as
\begin{equation} \label{eq:chairi}
    \textrm{CHAIR}_i = \frac{|\{\textrm{hallucinated objects}\}|}{|\{\textrm{all objects mentioned}\}|}
\end{equation}
as well as the sentence level $\textrm{CHAIR}_s$ to measure the proportion of hallucinated captions defined as
\begin{equation} \label{eq:chairs}
    \textrm{CHAIR}_s = \frac{|\{\textrm{captions with hallucinated objects}\}|}{|\{\textrm{all captions}\}|}.
\end{equation}

Since we tackle the problem of object hallucinations in generated captions $s$, we use the $\textrm{CHAIR}_i$ metric to extract the objects $\mathcal{O}_{s}=\{o_0,\dots,o_{z} \}$ and simultaneously label the extracted objects as either true, which is encoded as $0$, or hallucinated, which is encoded as $1$. For more details, see \cref{sec:chair_eval}.

\subsection{Large Vision Language Models} \label{subseq:eval_LVLMs}
We evaluate our approach on four SOTA open-source LVLMs, i.e., InstructBLIP (Vicuna-7B) \cite{Dai.2023b}, mPLUG-Owl (LLaMA-7B) \cite{Ye.2023}, MiniGPT-4 (Vicuna-7B) \cite{Zhu.2023}, and LLaVa 1.5 (Vicuna-7B) \cite{Huang.2023}, all of them using $G=32$ attention heads. We use nucleus sampling \cite{Holtzman.2020} and the prompt
\begin{center}
 \textit{"Describe all objects in the image."}   
\end{center}
for all image caption generations. For InstructBLIP and mPLUG-Owl, we create further image captions on MSCOCO using beam search \cite{Bisiani.1987}. \cref{tab:gen_config} states the detailed generation configurations of our experiments. The performance of the LVLMs considered with respect to the average number of generated objects per image as well as the hallucination rate in terms of $\textrm{CHAIR}_i$ (\cref{eq:chairi}) is summarized in \cref{tab:modeloverview}. Note that these results highly depend on the respective generation configurations, the underlying LLM (7B vs. 13B), as well as the chosen prompt \cite{Li.2023,Wang.2023}.

\subsection{Evaluation Metrics and $\quad\quad\quad$ Meta Models}
We evaluate our method based on the accuracy $ACC$, the area under receiver operator characteristic curve $AUROC$ \cite{Davis.2006} and the area under precision recall curve $AUPRC$ \cite{Davis.2006}. The receiver operator characteristic curve illustrates the performance of a binary classifier by plotting the true positive rate against the false positive rate at various decision thresholds indicating the ability to distinguish between both classes. The precision recall curve plots precision values against the recall at various decision thresholds accounting for imbalance in the underlying dataset. Since we observe imbalanced data with respect to object instance hallucinations (see \cref{tab:modeloverview}), the main focus in our evaluation is on the $AUROC$ and $AUPRC$ value. We compare two binary meta models, i.e., a classifier based on a logistic regression (LR) and a gradient boosting (GB) meta model.

\subsection{Baseline}
We use the reference-free token-level algebraic confidence TLC-A \cite{Petryk2023SimpleTC} as our baseline. We consider the log probability-based token-level confidence $L$ (see \cref{eq:logp}) and the entropy-based confidence $E$ (see \cref{eq:entropy}). For both confidence measures, we train a baseline classifier in the one-dimensional space, i.e.,
\begin{equation}
    f^{\mathrm{baseline}}:\mathbb{R} \rightarrow \{0,1\}
\end{equation}
with training set $\{(L_{o_j},y_{o_j}) |  j=0,\dots, |\mathcal{O}_{\mathcal{S}^{\textrm{train}}}|-1\}$ and $\{(E_{o_j},y_{o_j}) |  j=0,\dots, |\mathcal{O}_{\mathcal{S}^{\textrm{train}}}|-1\}$, respectively.

Note that a direct comparison of our approach to the detection methods listed in \cref{tab:related_work} is not possible. While our method tackles the problem of token-level object hallucination, the listed methods either evaluate hallucinations on a sentence- or subsentence-level on their own human-labeled dataset \cite{Gunjal.2023,Wang.2023} or based on atomic facts extracted from the image captions using LLMs, which are then labeled using LVLMs \cite{Wu.2024,Yin.2023,Jing.2023,Chen.2024}. Thus, the proposed methods \cite{Gunjal.2023} and \cite{Wang.2023} do not provide any information on which specific word of the respective sentence or subsentence is hallucinated. Similarly, \cite{Wu.2024,Yin.2023,Jing.2023,Chen.2024} are based on LLM-generated atomic facts which neither allow for a token-level evaluation and are not reproducible. Moreover, the analysis in \cite{Jing.2023} shows that the LLM-based atomic fact extraction already induces errors propagating through the detection and evaluation pipeline. To overcome these issues, we rely on the automated and reproducible CHAIR evaluation method (see \cref{subseq:haleval}).

\begin{table}[t!]
  \centering
  \caption{\textbf{Computational Time.} The average time for feature calculation per image (feature), classifier training using $4.000$ image captions (train) and inference on $1.000$ image captions (predict).}
  
   \resizebox{.9\columnwidth}{!}{
   \begin{tabular}{cccc}
          & feature (sec.) & train (sec.) & predict (sec.) \\
          \hline
          LR & $0.07174$ & $0.47187$ & $0.00198$ \\
          GB & $0.07174$ & $54.00010$ & $0.00385$ 
    \end{tabular}
    }

  \label{tab:comp_time} 
\end{table}

\begin{table*}[t!]
  \centering
  \caption{\textbf{Experimental Results based on Nucleus Sampling \cite{Holtzman.2020}.} Hallucination detection results on four SOTA LVLMs. \colorbox{textcolortab!20}{Ours} refers to the feature set $\mathcal{M}$. The best results in each block are highlighted.}
  
   \resizebox{2\columnwidth}{!}{
   \begin{tabular}{cc|cc|cc|cc}
        \toprule
        \multicolumn{8}{c}{MSCOCO \cite{Lin.2014}} \\
        \multicolumn{2}{c}{} & \multicolumn{2}{c}{$ACC \hspace{1.5mm} (\textrm{in} \hspace{1.5mm} \%) \hspace{1mm} \uparrow$} & \multicolumn{2}{c}{$AUROC \hspace{1.5mm} (\textrm{in} \hspace{1.5mm} \%) \hspace{1mm} \uparrow$} & \multicolumn{2}{c}{$AUPRC \hspace{1.5mm} (\textrm{in} \hspace{1.5mm} \%) \hspace{1mm} \uparrow$} \\
        && LR & GB & LR & GB & LR & GB \\
        \hline
        \multirow{3}{*}{InstructBLIP} & $L$ & 
        $89.46 ^{(\pm 1.4\mathrm{e}-1)}$ & $89.46 ^{(\pm 1.3\mathrm{e}-1)}$ & 
        $73.51 ^{(\pm 8.7\mathrm{e}-1)}$ & $73.16 ^{(\pm 9.0\mathrm{e}-1)}$ & 
        $27.07 ^{(\pm 2.1\mathrm{e}-0)}$ & $25.6 ^{(\pm 2.5\mathrm{e}-0)}$ \\
        & $E$ & 
        $89.49 ^{(\pm 1.6\mathrm{e}-1)}$ & $89.48 ^{(\pm 1.6\mathrm{e}-1)}$ & 
        $65.49 ^{(\pm 1.3\mathrm{e}-0)}$ & $66.23 ^{(\pm 1.5\mathrm{e}-0)}$ & 
        $15.38 ^{(\pm 5.7\mathrm{e}-1)}$ & $17.68 ^{(\pm 7.0\mathrm{e}-1)}$ \\
        \rowcolor{textcolortab!20}\cellcolor{white} & Ours & 
        $91.34 ^{(\pm 1.7\mathrm{e}-1)}$ & $\textbf{91.49} ^{(\pm 1.8\mathrm{e}-1)}$ & 
        $\textbf{89.93} ^{(\pm 8.9\mathrm{e}-1)}$ & $\textbf{89.93} ^{(\pm 7.3\mathrm{e}-1)}$ & 
        $56.07 ^{(\pm 1.2\mathrm{e}-0)}$ & $\textbf{56.71} ^{(\pm 7.6\mathrm{e}-0)}$ \\
        \midrule
        \multirow{3}{*}{mPLUG-Owl} & $L$ & 
        $72.42 ^{(\pm 4.3\mathrm{e}-1)}$ & $72.48 ^{(\pm 4.6\mathrm{e}-1)}$ & 
        $71.75 ^{(\pm 9.4\mathrm{e}-1)}$ & $71.86 ^{(\pm 9.3\mathrm{e}-1)}$ & 
        $51.21 ^{(\pm 1.2\mathrm{e}-0)}$ & $50.65 ^{(\pm 1.1\mathrm{e}-0)}$ \\
        & $E$ & 
        $70.06 ^{(\pm 4.9\mathrm{e}-1)}$ & $70.77 ^{(\pm 2.9\mathrm{e}-1)}$ & 
        $66.01 ^{(\pm 6.3\mathrm{e}-1)}$ & $68.33 ^{(\pm 6.1\mathrm{e}-1)}$ & 
        $40.09 ^{(\pm 8.2\mathrm{e}-1)}$ & $45.54 ^{(\pm 1.2\mathrm{e}-0)}$ \\
        \rowcolor{textcolortab!20}\cellcolor{white} & Ours & 
        $82.90 ^{(\pm 1.9\mathrm{e}-1)}$ & $\textbf{83.26} ^{(\pm 2.6\mathrm{e}-1)}$ & 
        $88.41 ^{(\pm 3.9\mathrm{e}-1)}$ & $\textbf{88.90} ^{(\pm 2.8\mathrm{e}-1)}$ & 
        $75.94 ^{(\pm 6.2\mathrm{e}-1)}$ & $\textbf{77.04} ^{(\pm 5.8\mathrm{e}-1)}$ \\
        \midrule
        \multirow{3}{*}{MiniGPT-4} & $L$ & 
        $86.91 ^{(\pm 3.6\mathrm{e}-1)}$ & $86.85 ^{(\pm 3.9\mathrm{e}-1)}$ & 
        $67.26 ^{(\pm 2.1\mathrm{e}-0)}$ & $67.01 ^{(\pm 2.1\mathrm{e}-0)}$ & 
        $26.25 ^{(\pm 1.7\mathrm{e}-0)}$ & $25.41 ^{(\pm 1.2\mathrm{e}-0)}$ \\
        & $E$ & 
        $86.84 ^{(\pm 3.6\mathrm{e}-1)}$ & $86.82 ^{(\pm 3.6\mathrm{e}-1)}$ & 
        $60.78 ^{(\pm 1.8\mathrm{e}-0)}$ & $63.19 ^{(\pm 1.2\mathrm{e}-0)}$ & 
        $15.77 ^{(\pm 6.7\mathrm{e}-1)}$ & $18.98 ^{(\pm 1.3\mathrm{e}-0)}$ \\
        \rowcolor{textcolortab!20}\cellcolor{white} & Ours & 
        $88.92 ^{(\pm 3.5\mathrm{e}-1)}$ & $\textbf{89.27} ^{(\pm 4.9\mathrm{e}-1)}$ & 
        $88.16 ^{(\pm 1.5\mathrm{e}-0)}$ & $\textbf{89.74} ^{(\pm 1.3\mathrm{e}-0)}$ & 
        $54.90 ^{(\pm 6.5\mathrm{e}-0)}$ & $\textbf{57.25} ^{(\pm 5.7\mathrm{e}-0)}$ \\
        \midrule
        \multirow{3}{*}{LLaVa} & $L$ & 
        $81.57 ^{(\pm 1.4\mathrm{e}-1)}$ & $81.49 ^{(\pm 1.6\mathrm{e}-1)}$ & 
        $70.53 ^{(\pm 8.7\mathrm{e}-1)}$ & $70.73 ^{(\pm 6.6\mathrm{e}-1)}$ & 
        $37.53 ^{(\pm 2.0\mathrm{e}-0)}$ & $36.59 ^{(\pm 1.7\mathrm{e}-0)}$ \\
        & $E$ & 
        $81.28 ^{(\pm 2.8\mathrm{e}-1)}$ & $81.26 ^{(\pm 2.9\mathrm{e}-1)}$ & 
        $62.73 ^{(\pm 9.0\mathrm{e}-1)}$ & $64.63 ^{(\pm 7.7\mathrm{e}-1)}$ & 
        $23.85 ^{(\pm 6.3\mathrm{e}-1)}$ & $27.52 ^{(\pm 4.6\mathrm{e}-1)}$ \\
        \rowcolor{textcolortab!20}\cellcolor{white} & Ours & 
        $87.25 ^{(\pm 2.0\mathrm{e}-1)}$ & $\textbf{87.78} ^{(\pm 3.0\mathrm{e}-1)}$ & 
        $90.05 ^{(\pm 4.0\mathrm{e}-1)}$ & $\textbf{91.01} ^{(\pm 4.3\mathrm{e}-1)}$ & 
        $70.15 ^{(\pm 1.0\mathrm{e}-0)}$ & $\textbf{72.58} ^{(\pm 1.3\mathrm{e}-0)}$ \\
        \hline
        
        \midrule  
        \multicolumn{8}{c}{BDD100K \cite{bdd100k}} \\
        \multicolumn{2}{c}{} & \multicolumn{2}{c}{$ACC \hspace{1.5mm} (\textrm{in} \hspace{1.5mm} \%) \hspace{1mm} \uparrow$} & \multicolumn{2}{c}{$AUROC \hspace{1.5mm} (\textrm{in} \hspace{1.5mm} \%) \hspace{1mm} \uparrow$} & \multicolumn{2}{c}{$AUPRC \hspace{1.5mm} (\textrm{in} \hspace{1.5mm} \%) \hspace{1mm} \uparrow$} \\
        && LR & GB & LR & GB & LR & GB \\
        \hline
        \multirow{3}{*}{InstructBLIP} & $L$ & 
        $77.54 ^{(\pm 3.3\mathrm{e}-1)}$ & $77.73 ^{(\pm 4.3\mathrm{e}-1)} $ & 
        $63.30 ^{(\pm 8.2\mathrm{e}-1)}$ & $63.63 ^{(\pm 1.1\mathrm{e}-0)} $ & 
        $31.80 ^{(\pm 1.1\mathrm{e}-0)}$ & $31.94 ^{(\pm 1.2\mathrm{e}-0)} $ \\
        & $E$ & 
        $77.86 ^{(\pm 3.6\mathrm{e}-1)}$ & $77.86 ^{(\pm 3.7\mathrm{e}-1)} $ & 
        $54.32 ^{(\pm 5.8\mathrm{e}-1)}$ & $56.71 ^{(\pm 6.3\mathrm{e}-1)} $ & 
        $23.28 ^{(\pm 7.6\mathrm{e}-1)}$ & $ 26.06 ^{(\pm 8.9\mathrm{e}-1)}$ \\
        \rowcolor{textcolortab!20}\cellcolor{white} & Ours & 
        $84.07 ^{(\pm 2.0\mathrm{e}-1)}$ & $ \textbf{84.40} ^{(\pm 3.1\mathrm{e}-1)}$ & 
        $87.75 ^{(\pm 2.8\mathrm{e}-1)}$ & $ \textbf{88.78} ^{(\pm 3.3\mathrm{e}-1)}$ & 
        $66.04 ^{(\pm 1.6\mathrm{e}-0)}$ & $ \textbf{69.55} ^{(\pm 2.0\mathrm{e}-0)}$ \\
        \midrule
        \multirow{3}{*}{mPLUG-Owl} & $L$ & 
        $66.52 ^{(\pm 4.4\mathrm{e}-1)}$ & $ 66.25 ^{(\pm 5.3\mathrm{e}-1)}$ & 
        $63.34 ^{(\pm 6.7\mathrm{e}-1)}$ & $ 63.40 ^{(\pm 6.4\mathrm{e}-1)}$ & 
        $44.44 ^{(\pm 1.1\mathrm{e}-0)}$ & $ 43.86 ^{(\pm 7.5\mathrm{e}-1)}$ \\
        & $E$ & 
        $66.42 ^{(\pm 4.8\mathrm{e}-1)}$ & $ 66.39 ^{(\pm 5.4\mathrm{e}-1)}$ & 
        $57.83 ^{(\pm 4.8\mathrm{e}-1)}$ & $ 60.54 ^{(\pm 2.9\mathrm{e}-1)}$ & 
        $37.20 ^{(\pm 5.4\mathrm{e}-1)}$ & $40.82 ^{(\pm 6.9\mathrm{e}-1)} $ \\
        \rowcolor{textcolortab!20}\cellcolor{white} & Ours & 
        $78.63 ^{(\pm 6.4\mathrm{e}-1)}$ & $\textbf{79.85} ^{(\pm 6.3\mathrm{e}-1)}$ & 
        $85.06 ^{(\pm 6.2\mathrm{e}-1)}$ & $\textbf{87.36} ^{(\pm 6.7\mathrm{e}-1)}$ & 
        $69.43 ^{(\pm 1.6\mathrm{e}-0)}$ & $\textbf{74.79} ^{(\pm 2.1\mathrm{e}-0)}$ \\
        \midrule
        \multirow{3}{*}{MiniGPT-4} & $L$ & 
        $70.92 ^{(\pm 3.3\mathrm{e}-1)}$ & $71.08 ^{(\pm 2.1\mathrm{e}-1)} $ & 
        $63.76 ^{(\pm 3.4\mathrm{e}-1)}$ & $63.44 ^{(\pm 4.4\mathrm{e}-1)} $ & 
        $37.47 ^{(\pm 5.2\mathrm{e}-1)}$ & $ 37.18 ^{(\pm 2.1\mathrm{e}-0)}$ \\
        & $E$ & 
        $71.30 ^{(\pm 2.7\mathrm{e}-1)}$ & $71.20 ^{(\pm 2.7\mathrm{e}-1)} $ & 
        $64.04 ^{(\pm 5.4\mathrm{e}-1)}$ & $ 63.47 ^{(\pm 5.9\mathrm{e}-1)}$ & 
        $38.03 ^{(\pm 5.0\mathrm{e}-1)}$ & $37.51 ^{(\pm 3.0\mathrm{e}-1)} $ \\
        \rowcolor{textcolortab!20}\cellcolor{white} & Ours & 
        $85.83 ^{(\pm 2.1\mathrm{e}-1)}$ & $\textbf{86.05} ^{(\pm 4.9\mathrm{e}-1)}$ & 
        $90.75 ^{(\pm 5.1\mathrm{e}-1)}$ & $\textbf{92.12} ^{(\pm 4.2\mathrm{e}-1)}$ & 
        $79.57 ^{(\pm 1.1\mathrm{e}-0)}$ & $\textbf{84.01} ^{(\pm 1.3\mathrm{e}-0)}$ \\
        \midrule
        \multirow{3}{*}{LLaVa} & $L$ & 
        $69.92 ^{(\pm 2.4\mathrm{e}-1)}$ & $70.02 ^{(\pm 3.0\mathrm{e}-1)} $ & 
        $61.25 ^{(\pm 9.4\mathrm{e}-1)}$ & $ 61.23 ^{(\pm 7.2\mathrm{e}-1)}$ & 
        $37.29 ^{(\pm 1.2\mathrm{e}-0)}$ & $37.03 ^{(\pm 1.1\mathrm{e}-0)} $ \\
        & $E$ & 
        $70.05 ^{(\pm 2.8\mathrm{e}-1)}$ & $ 70.03 ^{(\pm 2.8\mathrm{e}-1)}$ & 
        $56.57 ^{(\pm 4.5\mathrm{e}-1)}$ & $57.88 ^{(\pm 6.2\mathrm{e}-1)} $ & 
        $32.58 ^{(\pm 7.2\mathrm{e}-1)}$ & $34.65 ^{(\pm 7.6\mathrm{e}-1)} $ \\
        \rowcolor{textcolortab!20}\cellcolor{white} & Ours & 
        $80.93 ^{(\pm 4.6\mathrm{e}-1)}$ & $\textbf{82.39} ^{(\pm 3.3\mathrm{e}-1)}$ & 
        $87.46 ^{(\pm 3.8\mathrm{e}-1)}$ & $\textbf{89.56} ^{(\pm 2.1\mathrm{e}-1)}$ & 
        $70.39 ^{(\pm 1.5\mathrm{e}-0)}$ & $\textbf{76.59} ^{(\pm 8.0\mathrm{e}-1)}$ \\
        \bottomrule    
    \end{tabular}
    }

  \label{tab:results_sampling} 
\end{table*}

\section{\uppercase{Results}}
\label{sec:results}
\subsection{Hallucination Detection} \label{subsec:haldetect}

In this section, we discuss the performance of our proposed method on four SOTA LVLMs. We evaluate our method based on an LR and GB meta model (see \cref{sec:meta_config} for the configuration details). \cref{tab:comp_time} summarizes the computational time for calculating the input features (\cref{subsec:metrics}), training the meta model and the prediction time for hallucination detection. As shown in \cref{tab:ece}, both models provide a calibrated classification between true and hallucinated objects reflected by a small expected calibration error ($ECE$) \cite{PakdamanNaeini.2015}. \cref{tab:results_sampling} summarizes our detection results. We achieve an $ACC$ of up to $91.49 \%$, $AUROC$ values of up to $92.12 \%$, and $AUPRC$ values of up to $84.01 \%$ which clearly outperforms the TLC-A baselines $L$ and $E$ \cite{Petryk2023SimpleTC}. While the GB classifier outperforms the linear model for our method in all experiments, this result does not hold for the one-dimensional baselines $L$ and $E$. Especially for the log probability-based token-level confidence $L$, the linear model is superior to the GB classifier in most of the experiments.

Moreover, we observe better detection results with respect to $AUPRC$ on the BDD100K dataset than on the MSCOCO data. This behavior is expected since the MSCOCO dataset is widely used as an instruction tuning dataset for pre-trained LVLMs leading to lower hallucination rates on MSCOCO (see \cref{tab:modeloverview}). Thus, the LVLMs induce less positive (hallucinated) training samples when generating image captions, which makes the problem of learning the lightweight classifier $f$ more challenging. Simultaneously, we achieve higher $ACC$ values on the MSCOCO dataset indicating the insufficiency of the $ACC$ as an evaluation metric for imbalanced datasets. While, for the sake of completeness, we state the performance of our method with respect to the $ACC$, we emphasize the superior interpretability of the $AUROC$ and $AUPRC$ values for imbalanced datasets \cite{Davis.2006}.

\begin{table*}[t!]
  \centering
  \caption{\textbf{Experimental Results based on Beam Search \cite{Bisiani.1987}.} Hallucination detection results on SOTA LVLMs. \colorbox{textcolortab!20}{Ours} refers to the feature set $\mathcal{M}$. The best results in each block are highlighted.}
  
   \resizebox{2\columnwidth}{!}{
   \begin{tabular}{cc|cc|cc|cc}
        \toprule
        \multicolumn{8}{c}{MSCOCO \cite{Lin.2014}} \\
        \multicolumn{2}{c}{} & \multicolumn{2}{c}{$ACC \hspace{1.5mm} (\textrm{in} \hspace{1.5mm} \%) \hspace{1mm} \uparrow$} & \multicolumn{2}{c}{$AUROC \hspace{1.5mm} (\textrm{in} \hspace{1.5mm} \%) \hspace{1mm} \uparrow$} & \multicolumn{2}{c}{$AUPRC \hspace{1.5mm} (\textrm{in} \hspace{1.5mm} \%) \hspace{1mm} \uparrow$} \\
        && LR & GB & LR & GB & LR & GB \\
        \hline
        \multirow{3}{*}{InstructBLIP} & $L$ & 
        $92.27 ^{(\pm 1.6\mathrm{e}-1)}$ & $92.19 ^{(\pm 1.7\mathrm{e}-1)}$ & 
        $70.68 ^{(\pm 2.2e+0)}$ & $70.37 ^{(\pm 1.8e+0)}$ & 
        $21.57 ^{(\pm 2.9e+0)}$ & $19.56 ^{(\pm 2.5e+0)}$ \\
        & $E$ & 
        $92.30 ^{(\pm 1.7\mathrm{e}-1)}$ & $92.27 ^{(\pm 1.7\mathrm{e}-1)}$ & 
        $71.25 ^{(\pm 1.7e+0)}$ & $71.38 ^{(\pm 1.8e+0)}$ & 
        $23.69 ^{(\pm 3.5e+0)}$ & $22.05 ^{(\pm 2.9e+0)}$ \\
        \rowcolor{textcolortab!20}\cellcolor{white} & Ours & 
        $\textbf{93.07} ^{(\pm 1.3\mathrm{e}-1)}$ & $93.05 ^{(\pm 1.4\mathrm{e}-1)}$ & 
        $89.69 ^{(\pm 3.8\mathrm{e}-1)}$ & $\textbf{89.74} ^{(\pm 5.9\mathrm{e}-1)}$ & 
        $\textbf{47.45} ^{(\pm 4.9e+0)}$ & $47.40 ^{(\pm 5.0e+0)}$ \\
        \midrule
        \multirow{3}{*}{mPLUG-Owl} & $L$ & 
        $77.64 ^{(\pm 2.8\mathrm{e}-1)}$ & $77.61 ^{(\pm 3.5\mathrm{e}-1)}$ & 
        $72.00 ^{(\pm 5.6\mathrm{e}-1)}$ & $72.71 ^{(\pm 4.0\mathrm{e}-1)}$ & 
        $50.87 ^{(\pm 8.3\mathrm{e}-1)}$ & $50.35 ^{(\pm 8.4\mathrm{e}-1)}$ \\
        & $E$ & 
        $78.47 ^{(\pm 4.6\mathrm{e}-1)}$ & $78.47 ^{(\pm 4.4\mathrm{e}-1)}$ & 
        $72.58 ^{(\pm 7.5\mathrm{e}-1)}$ & $73.62 ^{(\pm 6.0\mathrm{e}-1)}$ & 
        $52.99 ^{(\pm 1.3e+0)}$ & $52.55 ^{(\pm 1.2e+0)}$ \\
        \rowcolor{textcolortab!20}\cellcolor{white} & Ours & 
        $84.10 ^{(\pm 3.4\mathrm{e}-1)}$ & $\textbf{84.57} ^{(\pm 4.5\mathrm{e}-1)}$ & 
        $88.85 ^{(\pm 3.4\mathrm{e}-1)}$ & $\textbf{89.54} ^{(\pm 3.7\mathrm{e}-1)}$ & 
        $71.89 ^{(\pm 9.6\mathrm{e}-1)}$ & $\textbf{73.03} ^{(\pm 1.1e+0)}$ \\
        \bottomrule    
    \end{tabular}
    }

  \label{tab:results_beam} 
\end{table*}

\begin{figure*}
\begin{subfigure}{.34\textwidth}
  \centering
  \includegraphics[width=\columnwidth]{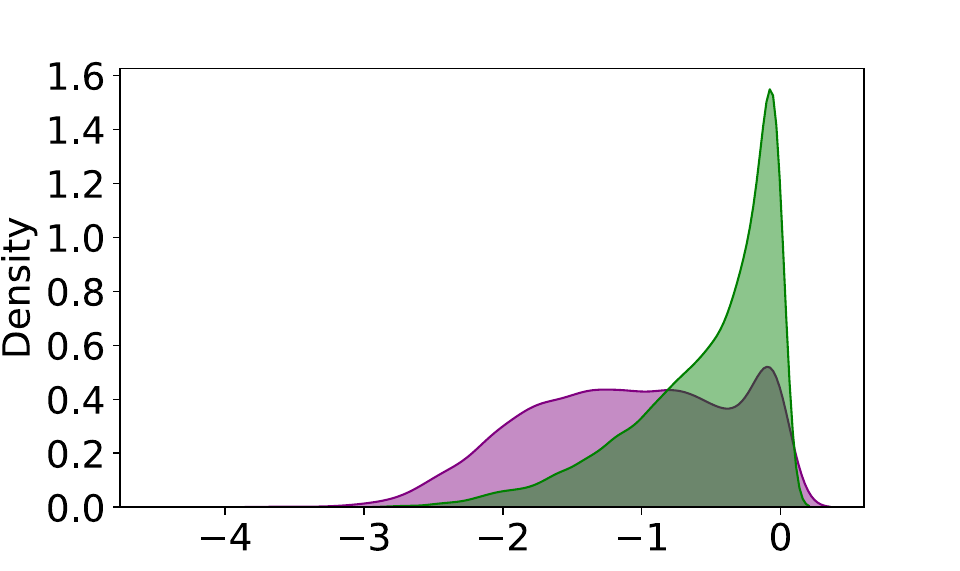}
  \caption{log probability $L$ (beam)}
  \label{fig:log_p_b}
\end{subfigure}%
\begin{subfigure}{.34\textwidth}
  \centering
  \includegraphics[width=\columnwidth]{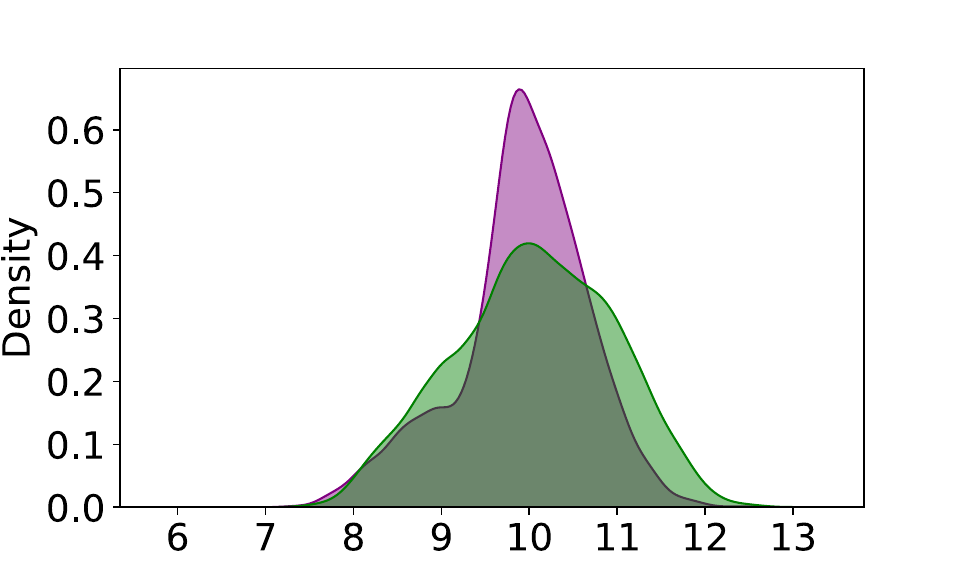}
  \caption{variance $V$ (beam)}
  \label{fig:var_b}
\end{subfigure}
\begin{subfigure}{.34\textwidth}
  \centering
  \includegraphics[width=\columnwidth]{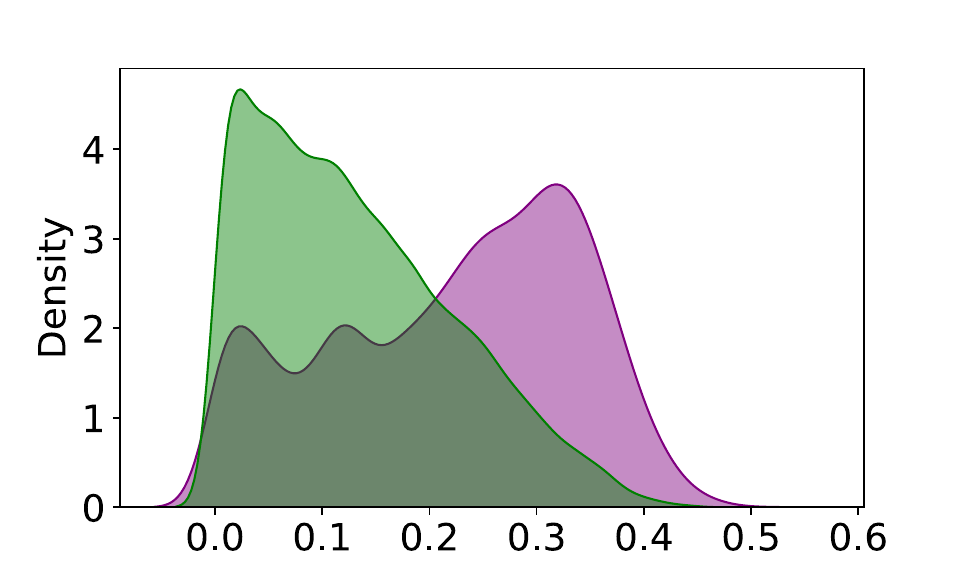}
  \caption{entropy $E$ (beam)}
  \label{fig:e_b}
\end{subfigure}%
\\
\begin{subfigure}{.34\textwidth}
  \centering
  \includegraphics[width=\columnwidth]{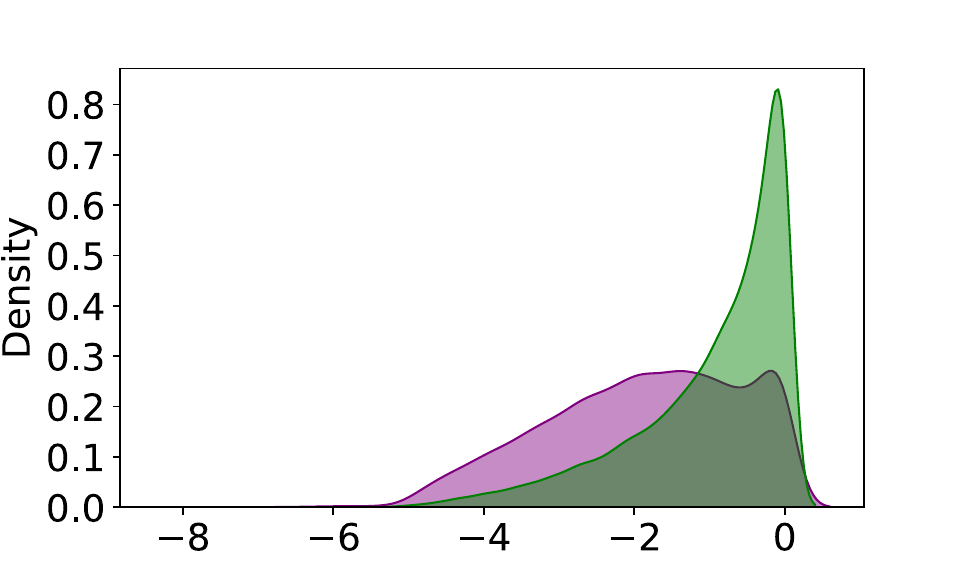}
  \caption{log probability $L$ (sample)}
  \label{fig:log_p_s}
\end{subfigure}
\begin{subfigure}{.34\textwidth}
  \centering
  \includegraphics[width=\columnwidth]{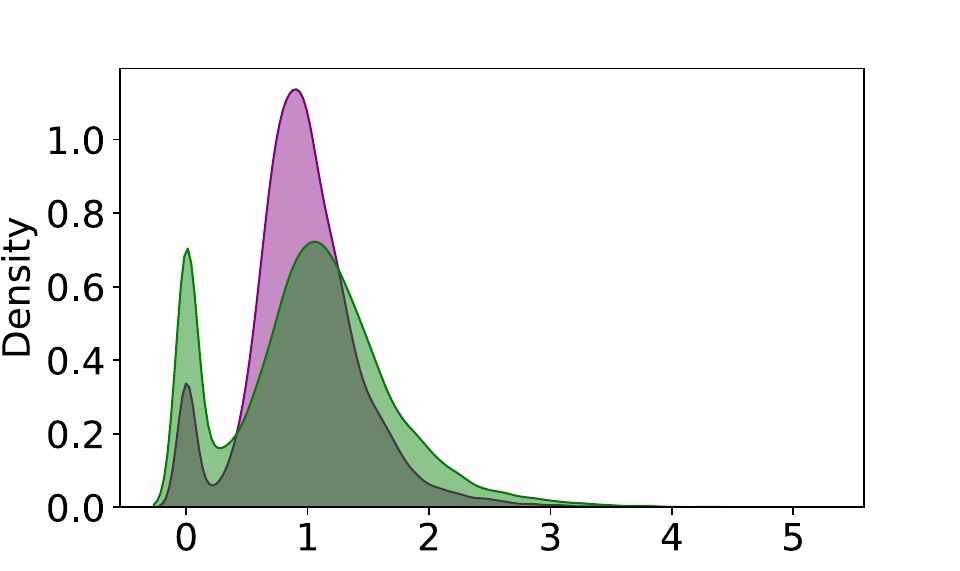}
  \caption{variance $V$ (sample)}
  \label{fig:var_s}
\end{subfigure}
\begin{subfigure}{.34\textwidth}
  \centering
  \includegraphics[width=\columnwidth]{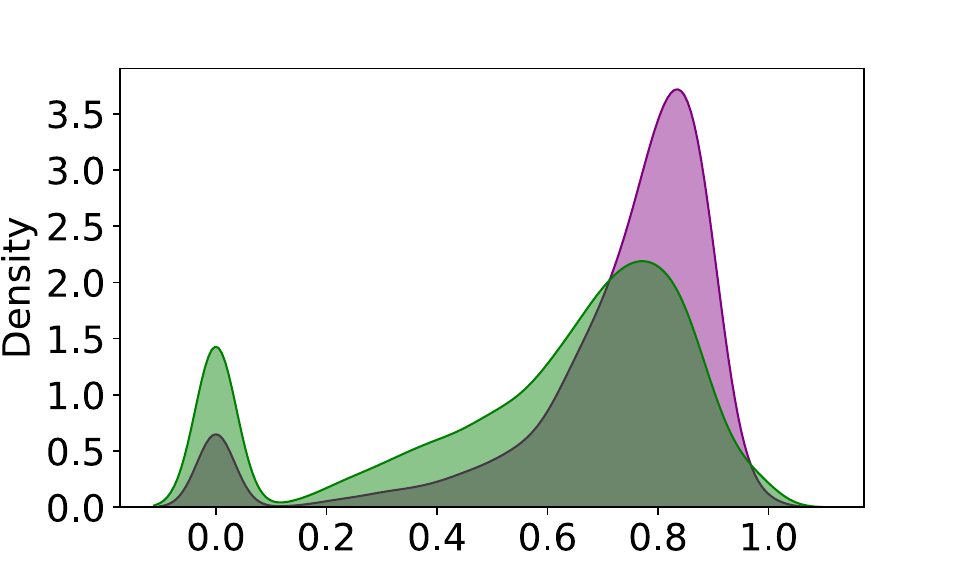}
  \caption{entropy $E$ (sample)}
  \label{fig:e_s}
\end{subfigure}
\caption{\textbf{Probability Features.} Visualization of the probability features log probability $L$ (\cref{eq:logp}), variance $V$ (\cref{eq:var}) and entropy $E$ (\cref{eq:entropy}) for  \colorbox{ao(english)!50}{true} and \colorbox{Mulberry!50}{hallucinated} objects. (beam) and (sample) reflect the beam search and nucleus sampling algorithm, respectively.}
\label{fig:beam_sample}
\end{figure*}

\subsection{Beam Search vs. Nucleus Sampling}
The results from the previous section also hold for image captions based on beam search. To compare both generation methods, we generate additional image captions for MSCOCO using the beam search algorithm for InstructBLIP \cite{Dai.2023b} and mPLUG-Owl \cite{Ye.2023}. The results are stated in \cref{tab:results_beam}. Again, generally the best results for our method are obtained by the gradient boosting classifier, clearly outperforming the single-feature baselines $L$ (\cref{eq:logp}) and $E$ (\cref{eq:entropy}). As we can see from \cref{tab:modeloverview}, beam search (beam) induces less hallucinations than nucleus sampling (sample) which is in line with previous findings \cite{Wang.2023,Li.2022}. Thus, the resulting classification problem (beam) is more challenging, i.e., in general we obtain better detection results for the sampling-based captions with respect to $AUROC$ and $AUPRC$ (see \cref{tab:results_sampling,tab:results_beam}).

While the log probability baseline $L$ (\cref{eq:logp}) shows similar classification performance for both, nucleus sampling- and beam search-based captions (see \cref{tab:results_sampling,tab:results_beam}), the entropy baseline $E$ (\cref{eq:entropy}) is more powerful for captions generated with beam search than for sampled descriptions. Even though we observe less hallucinations using the beam search algorithm (see \cref{tab:modeloverview}), we obtain better classification results from the entropy baseline $E$ (beam) than $E$ (sample). To this end, note that the entropy feature $E_{o_j}$ serves as a measure of uncertainty of generating token $t_{o_{j,s}}$ from the dictionary $\mathcal{V}$ (see \cref{sec:method}). Thus, we expect a higher entropy, i.e., a flatter probability distribution over the dictionary for hallucinated objects and a smaller entropy for true objects. In fact, we can see in \cref{fig:e_b,fig:e_s} that this holds for both, nucleus sampling and beam search. However, nucleus sampling induces generally smaller log probability values $L$ (sample) (see \cref{fig:log_p_b,fig:log_p_s}), and thus, a flatter probability distribution over the dictionary than beam search resulting in a smaller variance $V$ (sample) (see \cref{fig:var_b,fig:var_s}). Thus, for nucleus sampling, we obtain generally higher entropy values $E$ (sample) which results in a higher overlap of entropy values for true and hallucinated objects (see \cref{fig:e_s}), which makes the classification problem more challenging. For further insights we refer to \cref{sec:app_lasso}.

\begin{figure}[t!]
  \centering
  \includegraphics[width=\columnwidth]{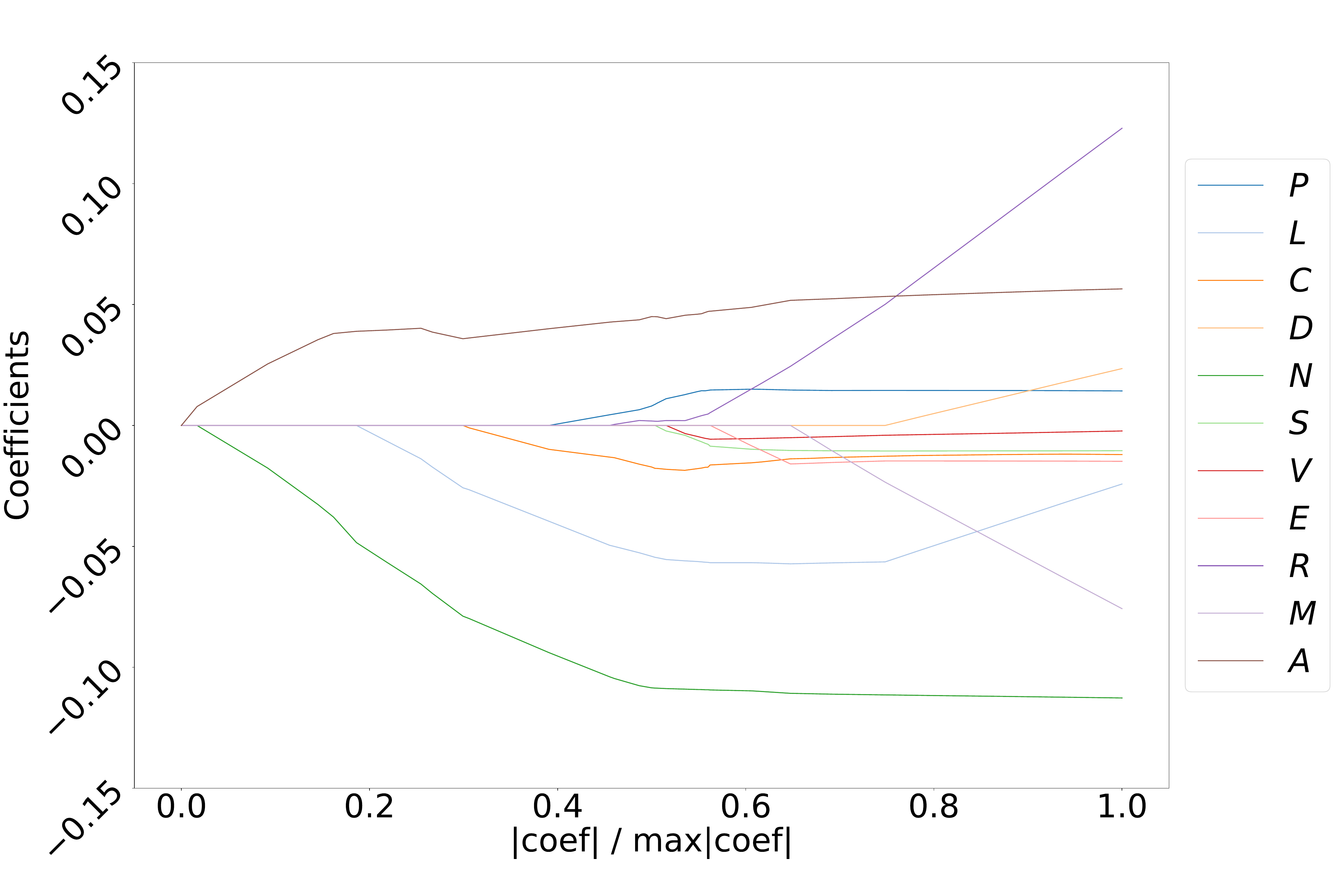}
  \caption{\textbf{LASSO Path.} LASSO path for $\mathcal{M}$. $A$ denotes the maximum of the absolute values of all $G$ weight coefficients for the attention features $A^{g}, g=0,\dots,G-1$.}
  \label{fig:lasso_mPLUGOwl_sampling} 
\end{figure}

\begin{table}[t!]
  \centering
  \caption{\textbf{Feature Rank.} The average rank of the features $\mathcal{M}$ in the LASSO paths of four SOTA art LVLMs.}
  
   \resizebox{\columnwidth}{!}{
   \begin{tabular}{ccl}
        \toprule
        avg. rank & feature & feature name \\
        \midrule    
        $1.375$ & $N$ & absolute occurrence (\cref{eq:absocc}) \\
        $2.125$ & $A$ & mean absolute attention (\cref{eq:att}) \\
        $3.625$ & $C$ & cumulated log probability (\cref{eq:cumlogp}) \\
        $5.125$ & $L$ & log probability (\cref{eq:logp}) \\       
        $6.875$ & $R$ & variation ratio (\cref{eq:varrat}) \\
        $6.875$ & $V$ & variance (\cref{eq:var}) \\
        $7.250$ & $E$ & entropy (\cref{eq:entropy}) \\
        $7.625$ & $D$ & probability difference (\cref{eq:probdiff}) \\
        $8.000$ & $S$ & score (\cref{eq:score}) \\
        $8.375$ & $M$ & probability margin (\cref{eq:probmargin}) \\
        $8.750$ & $P$ & relative position (\cref{eq:relpos}) \\        
        \bottomrule
    \end{tabular}
    }

  \label{tab:average_rank} 
\end{table}

\subsection{Feature Analysis}
\begin{figure}[t!]
  \centering
  \includegraphics[width=\columnwidth]{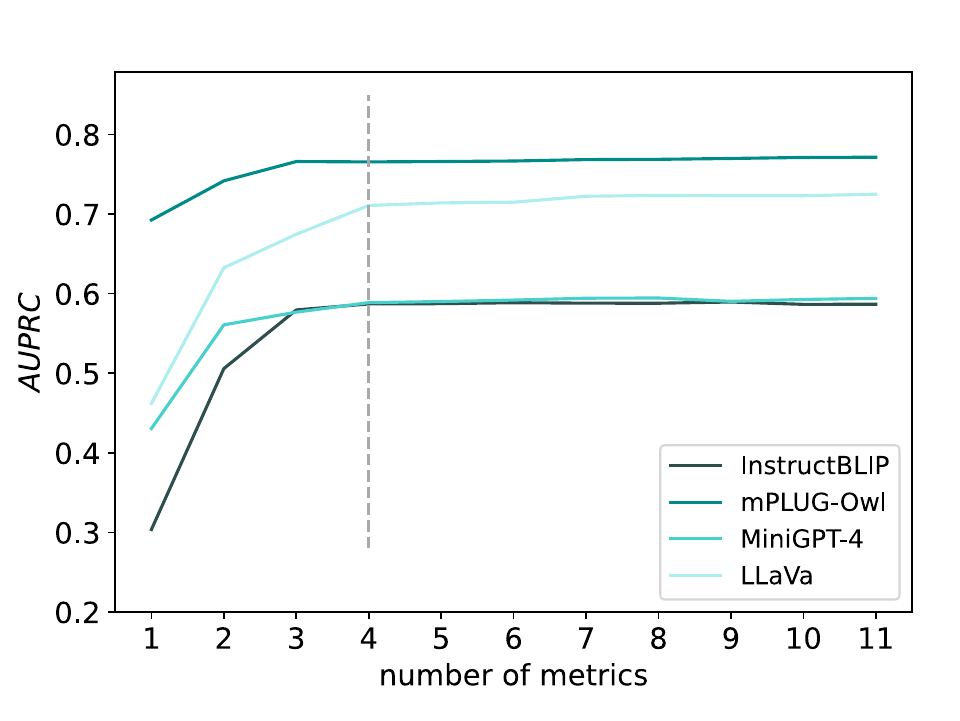}
  \caption{\textbf{$AUPRC$ as a Function of the Number of Features.} The classification performance of MetaToken in terms of $AUPRC$ as a function of the number of features for different LVLMs. The features are selected along the LASSO path of the respective LVLM.}
  \label{fig:auprc_metrics}
\end{figure}

In this section, we investigate the information contained in our proposed input features introduced in \cref{subsec:metrics}. A visualization of these statistics can be found in \cref{sec:app_metricvis}. We make use of the least absolute shrinkage and selection operator (LASSO) algorithm \cite{Efron.2004,Tibshirani.2018} to analyze the predictive power of the input features considered. The LASSO method performs a variable selection for a linear regression including the estimation of the corresponding coefficients ranking the most informative features. For the attention features (\cref{eq:att}), we use the maximum of the absolute values of all $G$ weight coefficients. \cref{fig:lasso_mPLUGOwl_sampling} shows the LASSO path for mPLUG-Owl. Our proposed attention features $A^{g}, g=0,\dots,G-1$ (\cref{eq:att}) are selected first, closely followed by the absolute occurrence $N$ (\cref{eq:absocc}), the log probability $L$ (\cref{eq:logp}) as well as the cumulated log probability $C$ (\cref{eq:cumlogp}). Moreover, the LASSO path indicates a minor relevance of the sequence score $S$ (\cref{eq:score}) and the variance $V$ (\cref{eq:var}) indicated through vanishing coefficients. We obtain similar results independently from the underlying LVLM or dataset (see \cref{sec:app_lasso}). \cref{tab:average_rank} lists the average rank of all features contained in the feature set $\mathcal{M}$ (\cref{eq:M}) during the LASSO selection. While most of the features are selected during the LASSO paths indicated through non-zero coefficients (see \cref{fig:lasso_full}), \cref{fig:auprc_metrics} shows that four features are usually enough to achieve high $AUPRC$ values. Further features only add minor additional information to the classifier.

Finally, we refer to \cref{sec:app_metricvis} to emphasize the importance of our statistical analysis based on the LASSO algorithm. While the relative position $P$ (\cref{eq:relpos}) and the probability margin $M$ (\cref{eq:probmargin}) might look like proper features to classify between hallucinated and true objects (see \cref{fig:metrics_full}), our analysis shows that these features only add minor information to the classifier reflected by an average rank of $8.750$ and $8.375$, respectively (see \cref{tab:average_rank}). 

\subsection{MetaToken and Revision of Image Descriptions}
\begin{table}[t!]
  \centering
  \renewcommand{\arraystretch}{.9}
  \caption{\textbf{Integration of MetaToken into LURE.} Results of MetaToken (Ours) plugged into the LURE mitigation method  \cite{Zhou.2023}. The superscripts denote the hallucination recall values for the respective method. PR and FPR denote the hallucination precision and false positive rate, respectively. The best results are highlighted, the second best results are underlined.}
  
\setlength{\tabcolsep}{.3em}
 \extrarowheight=\aboverulesep
    \addtolength{\extrarowheight}{\belowrulesep}
    \aboverulesep=0pt
    \belowrulesep=0pt
   \resizebox{\columnwidth}{!}{
   \begin{tabular}{clcccc}
        \toprule
        & Method & $\textrm{CHAIR}_i \downarrow$ & $\textrm{CHAIR}_s \downarrow$ & PR $\uparrow$ & FPR $\downarrow$ \\
        \hline
        \multirow{5}{*}{InstructBLIP} & $\textrm{LURE}^{76.5}$ & $10.9$ & $29.4$ & $17.5$ & $42.2$ \\
        & $\textrm{Ours}^{70}$ & $\textbf{9.8}$ & $\textbf{28.2}$ & $\textbf{46.5}$ & $ \textbf{9.4}$ \\
        & $\textrm{Ours}^{80}$ & $\underline{10.3}$ & $\underline{28.5}$ & $\underline{37.0}$ & $\underline{15.9}$ \\
        & $\textrm{Ours}^{90}$ & $10.6$ & $30.3$ & $26.3$ & $29.5$ \\
        & $\textrm{Ours}^{100}$ & $11.6$ & $29.4$ & $10.5$ & $100$ \\
        \midrule
        \multirow{5}{*}{mPLUG-Owl} & $\textrm{LURE}^{78.6}$ & $14.2$ & $37.1$ & $42.4$ & $46.4$ \\
        & $\textrm{Ours}^{70}$ & $14.4$ & $36.3$ & $\textbf{71.9}$ & $\textbf{11.9}$ \\
        & $\textrm{Ours}^{80}$ & $13.1$ & $33.5$ & $\underline{64.8}$ & $\underline{18.8}$ \\
        & $\textrm{Ours}^{90}$ & $\underline{12.2}$ & $\underline{31.1}$ & $55.1$ & $31.6$ \\
        & $\textrm{Ours}^{100}$ & $\textbf{12.1}$ & $\textbf{30.2}$ & $30.3$ & $100$ \\
        \bottomrule    
    \end{tabular}}

   \vspace{-4.5mm}
   \label{tab:lure}
\end{table}
In this section, we investigate MetaToken as a substitute for the LURE detection on the MSCOCO dataset. LURE \cite{Zhou.2023} serves as a hallucination mitigation method using a MiniGPT-4-based revisor to rectify image captions. To this end, LURE applies thresholds on the log probability $L$ (\cref{eq:logp}) and the relative position $P$ (\cref{eq:relpos}) to detect possible object hallucinations and replace them by the \emph{I-don't-know} string "[IDK]". The resulting image caption and the input image are fed into the revisor afterwards to rectify the detected tokens. In our experiments, we replace the threshold-based LURE detection with our proposed MetaToken method (see \cref{sec:method}).

First of all note that as in \cite{Zhao.2024}, we are not able to reproduce the results in \cite{Zhou.2023} with respect to InstructBLIP. While we achieve a hallucination rate of $10.4 \%$ for InstructBLIP captions (see \cref{tab:modeloverview}), the rectified image captions by LURE include $10.9 \%$ of hallucinations, even increasing the amount of hallucinated objects (see \cref{tab:lure}). We believe that this observation results from the fact that MiniGPT-4 has a higher hallucination rate than InstructBLIP (see \cref{tab:modeloverview}). Since the LURE detection has a false positive rate (FPR) of $42.2 \%$ (see \cref{tab:lure}), i.e., detects true objects as hallucinations, we guess that the MiniGPT-4-based revisor replaces true objects by hallucinated objects, even though the revisor is fine-tuned to mitigate hallucinations. However, applying our detection method, we are able to mitigate hallucinations achieving $\textrm{CHAIR}_i$ values of $9.8 \%$. To this end, note that we can control the precision-recall-ratio in our method by varying the decision threshold of our lightweight meta classifier. For a recall of $70 \%$, we observe a FPR of $9.4 \%$ only. Thus, we prevent the revisor from including additional hallucinations by replacing false positives, that is, true objects. The results in \cref{tab:lure} confirm our assumption on InstructBLIP: The higher the FPR, the higher the number of hallucinations induced by the revisor.

For mPLUG-Owl, LURE reduced the number of hallucinations by $52.98 \%$, i.e. from $30.2 \%$ to $14.2 \%$ with a FPR of $46.4 \%$. Since the revisor induces substantially less hallucinations than mPLUG-Owl, the correction of true positives outweighs the potential introduction of new hallucinations by replacing false positives. In fact, we observe from \cref{tab:lure} that higher recall values, and thus, higher FPRs, lead to consistently lower hallucination rates for mPLUG-Owl. However, note that the superior precision-recall-ratio from our approach, again, outperforms the LURE results: For a recall of $80 \%$ (which is closest to the LURE detection of $78.6 \%$), we reduce the proportion of hallucinations by $56.62 \%$ to $13.1 \%$, that is, $3.64 \%$ less hallucinations compared to the LURE baseline.

\section{\uppercase{Limitations}}
\label{sec:limitations}
Due to the lack of automated and reproducible token-level evaluation methods for attribute-level hallucinations, MetaToken is currently restricted to the problem of object hallucination detection, while the detection of attribute-level hallucinations remains an unsolved problem we will tackle in future work. Moreover, while the automated CHAIR method \cite{Rohrbach.2018} relies on ground truth labels, it still leads to mismatches due to misinterpretations of the generated language output of LVLMs.

\section{\uppercase{Conclusion}}
\label{sec:conclusion}
In this paper, we introduce MetaToken, a novel lightweight hallucination detection technique for LVLMs based on meta classification. Inspired by recently discovered causes of hallucinations, we propose and analyze a broad set of potential factors for hallucinations in LVLMs. Based on a comprehensive statistical analysis of these factors, we reveal key indicators of hallucinations. We evaluate our method on four SOTA LVLMs achieving $AUROC$ values of up to $92.12 \%$ and $AUPRC$ values of up to $84.01 \%$. Moreover, we show that our lightweight classifier detects hallucinations inducing an $ECE$ between $0.81 \%$ and $2.01 \%$. Finally, we demonstrate that MetaToken can be easily integrated into the LURE mitigation method reducing the hallucination rate by up to $56.62 \%$, i.e., $3.64 \%$ less hallucinations than the LURE baseline. As future work, we will tackle the problem of attribute-level hallucination detection for general visual question answering tasks.

\section*{\uppercase{Disclaimer}}
The results, opinions and conclusions expressed in this publication are not necessarily those of Volkswagen Aktiengesellschaft.

\bibliographystyle{apalike}
{\small
\bibliography{hal}}

\begin{thebibliography}{}

\bibitem[Anderson et~al., 2016]{Anderson.2016}
Anderson, P., Fernando, B., Johnson, M., and Gould, S. (2016).
\newblock Spice: Semantic propositional image caption evaluation.
\newblock In Leibe, B., Matas, J., Sebe, N., and Welling, M., editors, {\em Computer Vision -- ECCV 2016}, pages 382--398, Cham. {Springer International Publishing}.

\bibitem[Bisiani, 1987]{Bisiani.1987}
Bisiani, R. (1987).
\newblock Beam search.
\newblock {\em Encyclopedia of Artificial Intelligence}, pages 56--58.

\bibitem[Chen et~al., 2019]{Chen.2019}
Chen, T., Navratil, J., Iyengar, V., and Shanmugam, K. (2019).
\newblock Confidence scoring using whitebox meta-models with linear classifier probes.
\newblock In Chaudhuri, K. and Sugiyama, M., editors, {\em Proceedings of the Twenty-Second International Conference on Artificial Intelligence and Statistics}, volume~89 of {\em Proceedings of Machine Learning Research}, pages 1467--1475. PMLR.

\bibitem[Chen et~al., 2024]{Chen.2024}
Chen, X., Wang, C., Xue, Y., Zhang, N., Yang, X., Li, Q., Shen, Y., Liang, L., Gu, J., and Chen, H. (2024).
\newblock Unified hallucination detection for multimodal large language models.
\newblock In Ku, L.-W., Martins, A., and Srikumar, V., editors, {\em Proceedings of the 62nd Annual Meeting of the Association for Computational Linguistics (Volume 1: Long Papers)}, pages 3235--3252, Bangkok, Thailand. Association for Computational Linguistics.

\bibitem[Dai et~al., 2023a]{Dai.2023b}
Dai, W., Li, J., LI, D., Tiong, A., Zhao, J., Wang, W., Li, B., Fung, P.~N., and Hoi, S. (2023a).
\newblock Instructblip: Towards general-purpose vision-language models with instruction tuning.
\newblock In Oh, A., Neumann, T., Globerson, A., Saenko, K., Hardt, M., and Levine, S., editors, {\em Advances in Neural Information Processing Systems}, volume~36, pages 49250--49267. {Curran Associates, Inc}.

\bibitem[Dai et~al., 2023b]{Dai.2023}
Dai, W., Liu, Z., Ji, Z., Su, D., and Fung, P. (2023b).
\newblock Plausible may not be faithful: Probing object hallucination in vision-language pre-training.
\newblock In Vlachos, A. and Augenstein, I., editors, {\em Proceedings of the 17th Conference of the European Chapter of the Association for Computational Linguistics}, pages 2136--2148, Dubrovnik, Croatia. {Association for Computational Linguistics}.

\bibitem[Davis and Goadrich, 2006]{Davis.2006}
Davis, J. and Goadrich, M. (2006).
\newblock The relationship between precision-recall and roc curves.
\newblock In {\em Proceedings of the 23rd International Conference on Machine Learning}, ICML '06, pages 233--240, New York, NY, USA. {Association for Computing Machinery}.

\bibitem[Efron et~al., 2004]{Efron.2004}
Efron, B., Hastie, T., Johnstone, I., and Tibshirani, R. (2004).
\newblock Least angle regression.
\newblock {\em The Annals of Statistics}, 32(2):407--499.

\bibitem[Fieback et~al., 2023]{Fieback.2023}
Fieback, L., Dash, B., Spiegelberg, J., and Gottschalk, H. (2023).
\newblock Temporal performance prediction for~deep convolutional long short-term memory networks.
\newblock In Ifrim, G., Tavenard, R., Bagnall, A., Schaefer, P., Malinowski, S., Guyet, T., and Lemaire, V., editors, {\em Advanced Analytics and Learning on Temporal Data}, pages 145--158, Cham. {Springer Nature Switzerland}.

\bibitem[Fu et~al., 2023]{Fu.2023}
Fu, C., Chen, P., Shen, Y., Qin, Y., Zhang, M., Lin, X., Qiu, Z., Lin, W., Yang, J., Zheng, X., Li, K., Sun, X., and Ji, R. (2023).
\newblock Mme: A comprehensive evaluation benchmark for multimodal large language models.
\newblock {\em ArXiv}, abs/2306.13394.

\bibitem[Gao et~al., 2024]{Gao.2024}
Gao, H., Li, Y., Long, K., Yang, M., and Shen, Y. (2024).
\newblock A survey for foundation models in autonomous driving.
\newblock {\em ArXiv}, abs/2402.01105.

\bibitem[Gui et~al., 2024]{Gui.2024}
Gui, Y., Jin, Y., and Ren, Z. (2024).
\newblock Conformal alignment: Knowing when to trust foundation models with guarantees.
\newblock {\em ArXiv}.

\bibitem[Gunjal et~al., 2023]{Gunjal.2023}
Gunjal, A., Yin, J., and Bas, E. (2023).
\newblock Detecting and preventing hallucinations in large vision language models.
\newblock {\em ArXiv}, abs/2308.06394.

\bibitem[Hendrycks and Gimpel, 2017]{hendrycks17baseline}
Hendrycks, D. and Gimpel, K. (2017).
\newblock A baseline for detecting misclassified and out-of-distribution examples in neural networks.
\newblock In {\em Proceedings of International Conference on Learning Representations}.

\bibitem[Hessel et~al., 2021]{Hessel.2021}
Hessel, J., Holtzman, A., Forbes, M., {Le Bras}, R., and Choi, Y. (2021).
\newblock Clipscore: A reference-free evaluation metric for image captioning.
\newblock In Moens, M.-F., Huang, X., Specia, L., and Yih, S. W.-t., editors, {\em Proceedings of the 2021 Conference on Empirical Methods in Natural Language Processing}, pages 7514--7528, Online and Punta Cana, Dominican Republic. {Association for Computational Linguistics}.

\bibitem[Holtzman et~al., 2020]{Holtzman.2020}
Holtzman, A., Buys, J., {Du Li}, Forbes, M., and Choi, Y. (2020).
\newblock The curious case of neural text degeneration.
\newblock In {\em 8th International Conference on Learning Representations, ICLR 2020, Addis Ababa, Ethiopia, April 26-30, 2020}. OpenReview.net.

\bibitem[Huang et~al., 2023]{Huang.2023}
Huang, J., Zhang, J., Jiang, K., Qiu, H., and Lu, S. (2023).
\newblock Visual instruction tuning towards general-purpose multimodal model: A survey.
\newblock {\em ArXiv}, abs/2312.16602.

\bibitem[Jiang et~al., 2024]{Jiang.2024}
Jiang, Y., Omiye, J.~A., Zakka, C., Moor, M., Gui, H., Alipour, S., Mousavi, S.~S., Chen, J.~H., Rajpurkar, P., and Daneshjou, R. (2024).
\newblock Evaluating general vision-language models for clinical medicine.
\newblock {\em medRxiv}.

\bibitem[Jing et~al., 2023]{Jing.2023}
Jing, L., Li, R., Chen, Y., Jia, M., and {Du Xinya} (2023).
\newblock Faithscore: Evaluating hallucinations in large vision-language models.
\newblock {\em ArXiv}, abs/2311.01477.

\bibitem[Kowol et~al., 2020]{Kowol.2020}
Kowol, K., Rottmann, M., Bracke, S., and Gottschalk, H. (2020).
\newblock Yodar: Uncertainty-based sensor fusion for vehicle detection with camera and radar sensors.
\newblock In {\em International Conference on Agents and Artificial Intelligence}.

\bibitem[Lavie and Agarwal, 2007]{Lavie.2007}
Lavie, A. and Agarwal, A. (2007).
\newblock Meteor: An automatic metric for mt evaluation with high levels of correlation with human judgments.
\newblock In {\em Proceedings of the Second Workshop on Statistical Machine Translation}, StatMT '07, pages 228--231, USA. {Association for Computational Linguistics}.

\bibitem[Li et~al., 2023a]{Li.2023c}
Li, C., Wong, C., Zhang, S., Usuyama, N., Liu, H., Yang, J., Naumann, T., Poon, H., and Gao, J. (2023a).
\newblock Llava-med: Training a large language-and-vision assistant for biomedicine in one day.
\newblock In Oh, A., Neumann, T., Globerson, A., Saenko, K., Hardt, M., and Levine, S., editors, {\em Advances in Neural Information Processing Systems}, volume~36, pages 28541--28564. {Curran Associates, Inc}.

\bibitem[Li et~al., 2022]{Li.2022}
Li, J., LI, D., Xiong, C., and Hoi, S. (2022).
\newblock Blip: Bootstrapping language-image pre-training for unified vision-language understanding and generation.
\newblock In Chaudhuri, K., Jegelka, S., {Le Song}, Szepesvari, C., Niu, G., and Sabato, S., editors, {\em Proceedings of the 39th International Conference on Machine Learning}, volume 162 of {\em Proceedings of Machine Learning Research}, pages 12888--12900. PMLR.

\bibitem[Li et~al., 2023b]{Li.2023}
Li, Y., Du, Y., Zhou, K., Wang, J., Zhao, X., and Wen, J.-R. (2023b).
\newblock Evaluating object hallucination in large vision-language models.
\newblock In Bouamor, H., Pino, J., and Bali, K., editors, {\em Proceedings of the 2023 Conference on Empirical Methods in Natural Language Processing}, pages 292--305, Singapore. {Association for Computational Linguistics}.

\bibitem[Lin et~al., 2014]{Lin.2014}
Lin, T.-Y., Maire, M., Belongie, S., Hays, J., Perona, P., Ramanan, D., Doll{\'a}r, P., and Zitnick, C.~L. (2014).
\newblock Microsoft coco: Common objects in context.
\newblock In Fleet, D., Pajdla, T., Schiele, B., and Tuytelaars, T., editors, {\em Computer Vision -- ECCV 2014}, pages 740--755, Cham. {Springer International Publishing}.

\bibitem[Lin and Hauptmann, 2003]{Lin.2003}
Lin, W.-H. and Hauptmann, A. (2003).
\newblock Meta-classification: Combining multimodal classifiers.
\newblock In Za{\"i}ane, O.~R., Simoff, S.~J., and Djeraba, C., editors, {\em Mining Multimedia and Complex Data}, pages 217--231, Berlin, Heidelberg. {Springer Berlin Heidelberg}.

\bibitem[Liu et~al., 2023]{Liu.2023b}
Liu, F., Guan, T., Wu, X., Li, Z., Chen, L., Yacoob, Y., Manocha, D., and Zhou, T. (2023).
\newblock Hallusionbench: You see what you think? or you think what you see? an image-context reasoning benchmark challenging for gpt-4v(ision), llava-1.5, and other multi-modality models.
\newblock {\em ArXiv}, abs/2310.14566.

\bibitem[Liu et~al., 2024a]{liu2024mitigating}
Liu, F., Lin, K., Li, L., Wang, J., Yacoob, Y., and Wang, L. (2024a).
\newblock Mitigating hallucination in large multi-modal models via robust instruction tuning.
\newblock In {\em International Conference on Learning Representations}.

\bibitem[Liu et~al., 2024b]{Liu.2024}
Liu, H., Xue, W., Chen, Y., Chen, D., Zhao, X., Wang, K., Hou, L., Li, R.-Z., and Peng, W. (2024b).
\newblock A survey on hallucination in large vision-language models.
\newblock {\em ArXiv}, abs/2402.00253.

\bibitem[Liu et~al., 2022]{Liu.2022}
Liu, T., Zhang, Y., Brockett, C., Mao, Y., Sui, Z., Chen, W., and Dolan, B. (2022).
\newblock A token-level reference-free hallucination detection benchmark for free-form text generation.
\newblock In Muresan, S., Nakov, P., and Villavicencio, A., editors, {\em Proceedings of the 60th Annual Meeting of the Association for Computational Linguistics (Volume 1: Long Papers)}, pages 6723--6737, Dublin, Ireland. {Association for Computational Linguistics}.

\bibitem[Liu et~al., 2025]{Liu.2023}
Liu, Y., Duan, H., Zhang, Y., Li, B., Zhang, S., Zhao, W., Yuan, Y., Wang, J., He, C., Liu, Z., Chen, K., and Lin, D. (2025).
\newblock Mmbench: Is your multi-modal model an all-around player?
\newblock In Leonardis, A., Ricci, E., Roth, S., Russakovsky, O., Sattler, T., and Varol, G., editors, {\em Computer Vision -- ECCV 2024}, pages 216--233, Cham. Springer Nature Switzerland.

\bibitem[Lovenia et~al., 2024]{Lovenia.2023}
Lovenia, H., Dai, W., Cahyawijaya, S., Ji, Z., and Fung, P. (2024).
\newblock Negative object presence evaluation ({NOPE}) to measure object hallucination in vision-language models.
\newblock In Gu, J., Fu, T.-J.~R., Hudson, D., Celikyilmaz, A., and Wang, W., editors, {\em Proceedings of the 3rd Workshop on Advances in Language and Vision Research (ALVR)}, pages 37--58, Bangkok, Thailand. Association for Computational Linguistics.

\bibitem[Lu et~al., 2018]{Lu.2018}
Lu, J., Yang, J., Batra, D., and Parikh, D. (2018).
\newblock Neural baby talk.
\newblock In {\em 2018 IEEE/CVF Conference on Computer Vision and Pattern Recognition}, pages 7219--7228.

\bibitem[Maag et~al., 2020]{Maag.2020}
Maag, K., Rottmann, M., and Gottschalk, H. (2020).
\newblock Time-dynamic estimates of the reliability of deep semantic segmentation networks.
\newblock In {\em 2020 IEEE 32nd International Conference on Tools with Artificial Intelligence (ICTAI)}, pages 502--509.

\bibitem[Maag et~al., 2021]{Maag.2021}
Maag, K., Rottmann, M., Varghese, S., H{\"u}ger, F., Schlicht, P., and Gottschalk, H. (2021).
\newblock Improving video instance segmentation by light-weight temporal uncertainty estimates.
\newblock In {\em 2021 International Joint Conference on Neural Networks (IJCNN)}, pages 1--8.

\bibitem[{Pakdaman Naeini} et~al., 2015]{PakdamanNaeini.2015}
{Pakdaman Naeini}, M., Cooper, G., and Hauskrecht, M. (2015).
\newblock Obtaining well calibrated probabilities using bayesian binning.
\newblock {\em Proceedings of the AAAI Conference on Artificial Intelligence}, 29(1).

\bibitem[Papineni et~al., 2002]{Papineni.2002}
Papineni, K., Roukos, S., Ward, T., and Zhu, W.-J. (2002).
\newblock Bleu: A method for automatic evaluation of machine translation.
\newblock In Isabelle, P., Charniak, E., and Lin, D., editors, {\em Proceedings of the 40th Annual Meeting of the Association for Computational Linguistics}, pages 311--318, Philadelphia, Pennsylvania, USA. {Association for Computational Linguistics}.

\bibitem[Petryk et~al., 2023]{Petryk2023SimpleTC}
Petryk, S., Whitehead, S., Gonzalez, J., Darrell, T., Rohrbach, A., and Rohrbach, M. (2023).
\newblock Simple token-level confidence improves caption correctness.
\newblock {\em 2024 IEEE/CVF Winter Conference on Applications of Computer Vision (WACV)}, pages 5730--5740.

\bibitem[Radford et~al., 2021]{Radford.2021}
Radford, A., Kim, J.~W., Hallacy, C., Ramesh, A., Goh, G., Agarwal, S., Sastry, G., Askell, A., Mishkin, P., Clark, J., Krueger, G., and Sutskever, I. (2021).
\newblock Learning transferable visual models from natural language supervision.
\newblock In Meila, M. and Zhang, T., editors, {\em Proceedings of the 38th International Conference on Machine Learning}, volume 139 of {\em Proceedings of Machine Learning Research}, pages 8748--8763. PMLR.

\bibitem[Rohrbach et~al., 2018]{Rohrbach.2018}
Rohrbach, A., Hendricks, L.~A., Burns, K., Darrell, T., and Saenko, K. (2018).
\newblock Object hallucination in image captioning.
\newblock In Riloff, E., Chiang, D., Hockenmaier, J., and Tsujii, J., editors, {\em Proceedings of the 2018 Conference on Empirical Methods in Natural Language Processing}, pages 4035--4045, Brussels, Belgium. {Association for Computational Linguistics}.

\bibitem[Rottmann et~al., 2020]{Rottmann.2020}
Rottmann, M., Colling, P., {Paul Hack}, T., Chan, R., H{\"u}ger, F., Schlicht, P., and Gottschalk, H. (2020).
\newblock Prediction error meta classification in semantic segmentation: Detection via aggregated dispersion measures of softmax probabilities.
\newblock In {\em 2020 International Joint Conference on Neural Networks (IJCNN)}, pages 1--9.

\bibitem[Rottmann and Schubert, 2019]{Rottmann.2019}
Rottmann, M. and Schubert, M. (2019).
\newblock Uncertainty measures and prediction quality rating for the semantic segmentation of nested multi resolution street scene images.
\newblock In {\em 2019 IEEE/CVF Conference on Computer Vision and Pattern Recognition Workshops (CVPRW)}, pages 1361--1369.

\bibitem[Schubert et~al., 2021]{Schubert.2021}
Schubert, M., Kahl, K., and Rottmann, M. (2021).
\newblock Metadetect: Uncertainty quantification and prediction quality estimates for object detection.
\newblock In {\em 2021 International Joint Conference on Neural Networks (IJCNN)}, pages 1--10.

\bibitem[Shannon, 1948]{Shannon.1948}
Shannon, C.~E. (1948).
\newblock A mathematical theory of communication.
\newblock {\em The Bell System Technical Journal}, 27(3):379--423.

\bibitem[Tian et~al., 2024]{Tian.2024}
Tian, X., Gu, J., Li, B., Liu, Y., Hu, C., Wang, Y., Zhan, K., Jia, P., Lang, X., and Zhao, H. (2024).
\newblock Drivevlm: The convergence of autonomous driving and large vision-language models.
\newblock {\em ArXiv}, abs/2402.12289.

\bibitem[Tibshirani, 2018]{Tibshirani.2018}
Tibshirani, R. (2018).
\newblock Regression shrinkage and selection via the lasso.
\newblock {\em Journal of the Royal Statistical Society: Series B (Methodological)}, 58(1):267--288.

\bibitem[Vasudevan et~al., 2019]{Vasudevan.2019}
Vasudevan, V.~T., Sethy, A., and Ghias, A.~R. (2019).
\newblock Towards better confidence estimation for neural models.
\newblock In {\em ICASSP 2019 - 2019 IEEE International Conference on Acoustics, Speech and Signal Processing (ICASSP)}, pages 7335--7339.

\bibitem[Vedantam et~al., 2015]{Vedantam.2015}
Vedantam, R., Zitnick, C.~L., and Parikh, D. (2015).
\newblock Cider: Consensus-based image description evaluation.
\newblock In {\em 2015 IEEE Conference on Computer Vision and Pattern Recognition (CVPR)}, pages 4566--4575.

\bibitem[Wang et~al., 2023]{Wang.2023}
Wang, J., Zhou, Y., Xu, G., Shi, P., Zhao, C., Xu, H., Ye, Q., Yan, M., Zhang, J., Zhu, J., Sang, J., and Tang, H. (2023).
\newblock Evaluation and analysis of hallucination in large vision-language models.
\newblock {\em ArXiv}, abs/2308.15126.

\bibitem[Wang et~al., 2024]{Wang.2024}
Wang, L., He, J., Li, S., Liu, N., and Lim, E.-P. (2024).
\newblock Mitigating fine-grained hallucination by~fine-tuning large vision-language models with~caption rewrites.
\newblock In Rudinac, S., Hanjalic, A., Liem, C., Worring, M., J{\'o}nsson, B.~{\TH}., Liu, B., and Yamakata, Y., editors, {\em MultiMedia Modeling}, pages 32--45, Cham. {Springer Nature Switzerland}.

\bibitem[Wu et~al., 2024]{Wu.2024}
Wu, J., Liu, Q., Wang, D., Zhang, J., Wu, S., Wang, L., and Tan, T. (2024).
\newblock Logical closed loop: Uncovering object hallucinations in large vision-language models.
\newblock In Ku, L.-W., Martins, A., and Srikumar, V., editors, {\em Findings of the Association for Computational Linguistics ACL 2024}, pages 6944--6962, Bangkok, Thailand and virtual meeting. Association for Computational Linguistics.

\bibitem[Xing et~al., 2024]{Xing.2024}
Xing, S., Zhao, F., Wu, Z., An, T., Chen, W., Li, C., Zhang, J., and Dai, X. (2024).
\newblock Efuf: Efficient fine-grained unlearning framework for mitigating hallucinations in multimodal large language models.
\newblock {\em ArXiv}, abs/2402.09801.

\bibitem[Ye et~al., 2023]{Ye.2023}
Ye, Q., Xu, H., Xu, G., Ye, J., Yan, M., Zhou, Y., Wang, J., Hu, A., Shi, P., Shi, Y., Li, C., Xu, Y., Chen, H., Tian, J., Qi, Q., Zhang, J., and Huang, F. (2023).
\newblock mplug-owl: Modularization empowers large language models with multimodality.
\newblock {\em ArXiv}, abs/2304.14178.

\bibitem[Yin et~al., 2023]{Yin.2023}
Yin, S., Fu, C., Zhao, S., Xu, T., Wang, H., Sui, D., Shen, Y., Li, K., Sun, X., and Chen, E. (2023).
\newblock Woodpecker: Hallucination correction for multimodal large language models.
\newblock {\em ArXiv}, abs/2310.16045.

\bibitem[Yu et~al., 2020]{bdd100k}
Yu, F., Chen, H., Wang, X., Xian, W., Chen, Y., Liu, F., Madhavan, V., and Darrell, T. (2020).
\newblock Bdd100k: A diverse driving dataset for heterogeneous multitask learning.
\newblock In {\em IEEE/CVF Conference on Computer Vision and Pattern Recognition (CVPR)}.

\bibitem[Yu et~al., 2024]{Yu.2023}
Yu, W., Yang, Z., Li, L., Wang, J., Lin, K., Liu, Z., Wang, X., and Wang, L. (2024).
\newblock {MM}-vet: Evaluating large multimodal models for integrated capabilities.
\newblock In Salakhutdinov, R., Kolter, Z., Heller, K., Weller, A., Oliver, N., Scarlett, J., and Berkenkamp, F., editors, {\em Proceedings of the 41st International Conference on Machine Learning}, volume 235 of {\em Proceedings of Machine Learning Research}, pages 57730--57754. PMLR.

\bibitem[Zhao et~al., 2024]{Zhao.2024}
Zhao, L., Deng, Y., Zhang, W., and Gu, Q. (2024).
\newblock Mitigating object hallucination in large vision-language models via classifier-free guidance.
\newblock {\em ArXiv}, abs/2402.08680.

\bibitem[Zhou et~al., 2023]{Zhou.2023}
Zhou, Y., Cui, C., Yoon, J., Zhang, L., Deng, Z., Finn, C., Bansal, M., and Yao, H. (2023).
\newblock Analyzing and mitigating object hallucination in large vision-language models.
\newblock {\em ArXiv}, abs/2310.00754.

\bibitem[Zhu et~al., 2023]{Zhu.2023}
Zhu, D., Chen, J., Shen, X., Li, X., and Elhoseiny, M. (2023).
\newblock Minigpt-4: Enhancing vision-language understanding with advanced large language models.
\newblock {\em ArXiv}, abs/2304.10592.

\end{thebibliography}

\clearpage
\section{\uppercase{Appendix}}
\subsection{Configurations}
\subsubsection{LVLM Configurations}

The detailed configuration settings for the LVLM image caption generation used in our experiments can be found in \cref{tab:gen_config}. For those parameters not listed in \cref{tab:gen_config}, we refer to the default settings.

\subsubsection{Meta Classifier Configurations}  \label{sec:meta_config}
In our experiments, we use the sklearn.linear\_model.LogisticRegression\footnote{\url{https://scikit-learn.org/1.3/modules/generated/sklearn.linear_model.LogisticRegression.html}} and sklearn.ensemble.GradientBoostingClassifier\footnote{\url{https://scikit-learn.org/1.3/modules/generated/sklearn.ensemble.GradientBoostingClassifier.html}} methods with scikit-learn version $1.3.2$. \cref{tab:meta_config} states the applied configurations. For those parameters not listed in \cref{tab:meta_config}, we refer to the default settings.

\begin{table}[h!]
  \centering
  \caption{\textbf{LVLM Generation Configurations}. The generation configurations applied in our experiments for nucleus sampling \cite{Holtzman.2020} and beam search \cite{Bisiani.1987}, respectively.}
  
   \resizebox{1.0\columnwidth}{!}{
   \begin{tabular}{lcc}
        \toprule
        Parameter & Nucleus Sampling & Beam Search \\
        \midrule  
        $\textrm{do\_sample}$ & $\textrm{True}$ & $\textrm{False}$ \\
        $\textrm{top\_p}$ & $\textrm{0.9}$ & $\textrm{0.9}$ \\
        $\textrm{temperature}$ & $\textrm{1}$ & $\textrm{1}$ \\
        $\textrm{num\_beams}$ & $\textrm{1}$ & $\textrm{5}$ \\
        $\textrm{max\_length}$ & $\textrm{256}$ & $\textrm{256}$ \\
        $\textrm{min\_length}$ & $\textrm{1}$ & $\textrm{1}$ \\
        $\textrm{repetition\_penalty}$ & $\textrm{1}$ & $\textrm{1.5}$ \\
        $\textrm{length\_penalty}$ & $\textrm{1}$ & $\textrm{1}$ \\
        \bottomrule
    \end{tabular}
    }

  \label{tab:gen_config}
\end{table}

\begin{table}[h!]
  \centering
  \caption{\textbf{Classifier Configurations.} The configurations applied in our experiments for the sklearn.linear\_model.LogisticRegression (LR) and sklearn.ensemble.GradientBoostingClassifier (GB) classifier.}
  
   \resizebox{.6\columnwidth}{!}{
   \begin{tabular}{ccc}
        \toprule
        Parameter & LR & GB \\
        \hline
        scikit-learn & $1.3.2$ & $1.3.2$ \\
        random\_state & $0$ & $0$ \\
        solver & saga & - \\
        max\_iter & $1000$ & - \\
        tol & $1\mathrm{e}-3$ & - \\
        \bottomrule    
    \end{tabular}}

  \label{tab:meta_config} 
\end{table}

\subsection{CHAIR Evaluation} \label{sec:chair_eval}
Initially, the CHAIR metric \cite{Rohrbach.2018} has been introduced to measure hallucinations for the MSCOCO dataset \cite{Lin.2014}. CHAIR measures the proportion of hallucinated MSCOCO objects by matching the generated text against the ground truth objects provided in the MSCOCO image captioning and object detection datasets, and thus, provides a binary label for every generated MSCOCO object category and a wide range of corresponding synonyms \cite{Lu.2018} indicating whether the object is true, i.e., contained in the image, or hallucinated. We follow the same methodology to evaluate hallucinations on the BDD100K dataset \cite{bdd100k}. For every BDD100K object category, we create a comprehensive list of synonyms (see \cref{tab:bddsyn}) and match the LVLM output against the ground truth labels of the BDD100K object detection dataset.
\begin{table*}[h]
  \centering
  \caption{\textbf{BDD100K Synonyms.} A list of frequently used synonyms for the BDD100K object categories \cite{bdd100k}.}
  
   \begin{tabular}{ll}
        \toprule
        Object Category & Synonyms \\
        \hline
        person & human, man, woman, driver, people, someone, somebody, citizen, human being, walker, \\
        & pedestrian \\
        rider & cyclist, bicyclist, bike rider, biker, motorcyclist, motor biker, motorbike user \\
        & motorcycle user \\
        car & automobile, vehicle, auto, suv, motorcar, ride, roadster, taxi \\
        bus & coach, minibus, shuttle, omnibus, motorbus, passenger vehicle, trolleybus, school bus, \\
        & tour bus \\
        truck & lorry, pickup, van, semi-truck, rig, dump truck, cargo truck, delivery truck, garbage truck \\
        bike & bicycle, cycle, pedal bike, road bike, mountain bike, velocipede \\
        motor & motorcycle, scooter \\
        traffic light & stoplight, signal light, traffic signal, red light, green light, traffic control signal, road signal, \\
        & semaphore, stop light \\
        traffic sign & direction sign, railroad crossing sign, road sign, signpost, traffic marker, stop sign \\
        train & metro, tram \\
        \bottomrule    
    \end{tabular}

  \label{tab:bddsyn} 
\end{table*}

\subsection{Extended Features}
For the LVLMs considered which use the same vision encoder as CLIP \cite{Radford.2021}, that is, mPLUG-Owl \cite{Ye.2023} and LLaVa \cite{Huang.2023}, we evaluate the extended feature set $\mathcal{M_{o_j}}^{\textrm{clip}}$ including the \textbf{CLIPScore} \cite{Hessel.2021}
\begin{equation} \label{eq:clip}
    \textrm{CLIP}_{o_j} = \max(100*\cos(\mathcal{E}_{x}, \mathcal{E}_{{Cap}_{o_j}}),0)
\end{equation}
with the visual CLIP embedding $\mathcal{E}_{x}$ for image $x$ and the textual CLIP embedding $\mathcal{E}_{{Cap}_{o_j}}$ for the caption ${Cap}_{o_j}$ which is defined as \textit{"a photo of a <$o_j$>"}. This results in the extended set of features
\begin{equation} \label{eq:M_clip}
\begin{split}
    \mathcal{M}^{\mathrm{clip}}_{o_j} = \{P_{o_j}, N_{o_j}, A^{0}_{o_j},\dots, A^{G-1}_{o_j}, L_{o_j}, C_{o_j}, \\
    S_{o_j}, V_{o_j}, E_{o_j}, R_{o_j}, M_{o_j}, D_{o_j}, \textrm{CLIP}_{o_j}\}
\end{split}
\end{equation}
with cardinality $|\mathcal{M}^{\mathrm{clip}}_{o_j}| = 11+G$.

\begin{table*}[t!]
  \centering
  \caption{\textbf{Experimental Results based on the extended feature set $\mathcal{M}^{\mathrm{clip}}_{o_j}$.} \colorbox{textcolortab!20}{Ours} refers to the feature set $\mathcal{M}$ (\cref{eq:M}), while \colorbox{textcolortab!50}{$\textrm{Ours}^{\textrm{clip}}$} is based on the extended set $\mathcal{M}^{\textrm{clip}}$ (\cref{eq:M_clip}). The best results in each block are highlighted.}
  
   \resizebox{2\columnwidth}{!}{
   \begin{tabular}{cc|cc|cc|cc}
        \toprule
        \multicolumn{8}{c}{MSCOCO \cite{Lin.2014}} \\
        \multicolumn{2}{c}{} & \multicolumn{2}{c}{$ACC \hspace{1.5mm} (\textrm{in} \hspace{1.5mm} \%) \hspace{1mm} \uparrow$} & \multicolumn{2}{c}{$AUROC \hspace{1.5mm} (\textrm{in} \hspace{1.5mm} \%) \hspace{1mm} \uparrow$} & \multicolumn{2}{c}{$AUPRC \hspace{1.5mm} (\textrm{in} \hspace{1.5mm} \%) \hspace{1mm} \uparrow$} \\
        && LR & GB & LR & GB & LR & GB \\
        \hline
        & $L$ & 
        $72.42 ^{(\pm 4.3\mathrm{e}-1)}$ & $72.48 ^{(\pm 4.6\mathrm{e}-1)}$ & 
        $71.75 ^{(\pm 9.4\mathrm{e}-1)}$ & $71.86 ^{(\pm 9.3\mathrm{e}-1)}$ & 
        $51.21 ^{(\pm 1.2\mathrm{e}-0)}$ & $50.65 ^{(\pm 1.1\mathrm{e}-0)}$ \\
        & $E$ & 
        $70.06 ^{(\pm 4.9\mathrm{e}-1)}$ & $70.77 ^{(\pm 2.9\mathrm{e}-1)}$ & 
        $66.01 ^{(\pm 6.3\mathrm{e}-1)}$ & $68.33 ^{(\pm 6.1\mathrm{e}-1)}$ & 
        $40.09 ^{(\pm 8.2\mathrm{e}-1)}$ & $45.54 ^{(\pm 1.2\mathrm{e}-0)}$ \\
        \rowcolor{textcolortab!20}\cellcolor{white} & Ours & 
        $82.90 ^{(\pm 1.9\mathrm{e}-1)}$ & $\textbf{83.26} ^{(\pm 2.6\mathrm{e}-1)}$ & 
        $88.41 ^{(\pm 3.9\mathrm{e}-1)}$ & $\textbf{88.90} ^{(\pm 2.8\mathrm{e}-1)}$ & 
        $75.94 ^{(\pm 6.2\mathrm{e}-1)}$ & $\textbf{77.04} ^{(\pm 5.8\mathrm{e}-1)}$ \\
        \rowcolor{textcolortab!50}\cellcolor{white} \multirow{-4}{*}{mPLUG-Owl} & $\textrm{Ours}^{\mathrm{clip}}$ & 
        $84.80 ^{(\pm 3.2\mathrm{e}-1)}$ & $\textbf{85.01} ^{(\pm 3.6\mathrm{e}-1)}$ & 
        $90.91 ^{(\pm 3.1\mathrm{e}-1)}$ & $\textbf{91.12} ^{(\pm 2.3\mathrm{e}-1)}$ & 
        $80.28 ^{(\pm 4.9\mathrm{e}-1)}$ & $\textbf{80.78} ^{(\pm 4.5\mathrm{e}-1)}$ \\
        \midrule
        & $L$ & 
        $81.57 ^{(\pm 1.4\mathrm{e}-1)}$ & $81.49 ^{(\pm 1.6\mathrm{e}-1)}$ & 
        $70.53 ^{(\pm 8.7\mathrm{e}-1)}$ & $70.73 ^{(\pm 6.6\mathrm{e}-1)}$ & 
        $37.53 ^{(\pm 2.0\mathrm{e}-0)}$ & $36.59 ^{(\pm 1.7\mathrm{e}-0)}$ \\
        & $E$ & 
        $81.28 ^{(\pm 2.8\mathrm{e}-1)}$ & $81.26 ^{(\pm 2.9\mathrm{e}-1)}$ & 
        $62.73 ^{(\pm 9.0\mathrm{e}-1)}$ & $64.63 ^{(\pm 7.7\mathrm{e}-1)}$ & 
        $23.85 ^{(\pm 6.3\mathrm{e}-1)}$ & $27.52 ^{(\pm 4.6\mathrm{e}-1)}$ \\
        \rowcolor{textcolortab!20}\cellcolor{white} & Ours & 
        $87.25 ^{(\pm 2.0\mathrm{e}-1)}$ & $\textbf{87.78} ^{(\pm 3.0\mathrm{e}-1)}$ & 
        $90.05 ^{(\pm 4.0\mathrm{e}-1)}$ & $\textbf{91.01} ^{(\pm 4.3\mathrm{e}-1)}$ & 
        $70.15 ^{(\pm 1.0\mathrm{e}-0)}$ & $\textbf{72.58} ^{(\pm 1.3\mathrm{e}-0)}$ \\
        \rowcolor{textcolortab!50}\cellcolor{white} \multirow{-4}{*}{LLaVa} & $\textrm{Ours}^{\mathrm{clip}}$ & 
        $87.93 ^{(\pm 1.2\mathrm{e}-1)}$ & $\textbf{88.36} ^{(\pm 1.6\mathrm{e}-1)}$ & 
        $91.73 ^{(\pm 2.4\mathrm{e}-1)}$ & $\textbf{92.50} ^{(\pm 2.5\mathrm{e}-1)}$ & 
        $73.00 ^{(\pm 7.8\mathrm{e}-1)}$ & $\textbf{75.08} ^{(\pm 8.5\mathrm{e}-1)}$ \\
        \bottomrule    
    \end{tabular}
    }

  \label{tab:results_clip} 
\end{table*}

The detection results using the extended feature set $\mathcal{M}^{\mathrm{clip}}_{o_j}$ can be found in \cref{tab:results_clip}. The additional CLIPScore feature (\cref{eq:clip}) improves the $ACC$ by up to $1.90 pp$, the $AUROC$ by up to $2.50 pp$ and $AUPRC$ by up to $4.34 pp$ underlining the informative content provided by the cosine similarity between the image and text embedding. However, with $\mathcal{C}=\{c_{0}, \dots , c_{79} \}$ denoting the set of all $80$ MSCOCO objects, note that the CLIPScore feature can only be calculated in a lightweight manner if the objects of interest $\mathcal{C}$ are known in advance (to calculate the text embedding $\mathcal{E}_{{Cap}_{o_j}} \textrm{for all possible objects} \hspace{1mm} o_j \in \mathcal{C}$ \textrm{beforehand}) and if the CLIP image embedding $\mathcal{E}_{x}$ is calculated during the generation process. Since our focus is on a general method without any restrictions, we stick to the more general feature set $\mathcal{M}$ (\cref{eq:M}) in this paper.

\subsection{LASSO Path} \label{sec:app_lasso}

\begin{figure*}
\begin{subfigure}{.5\textwidth}
  \centering
  \includegraphics[width=\textwidth]{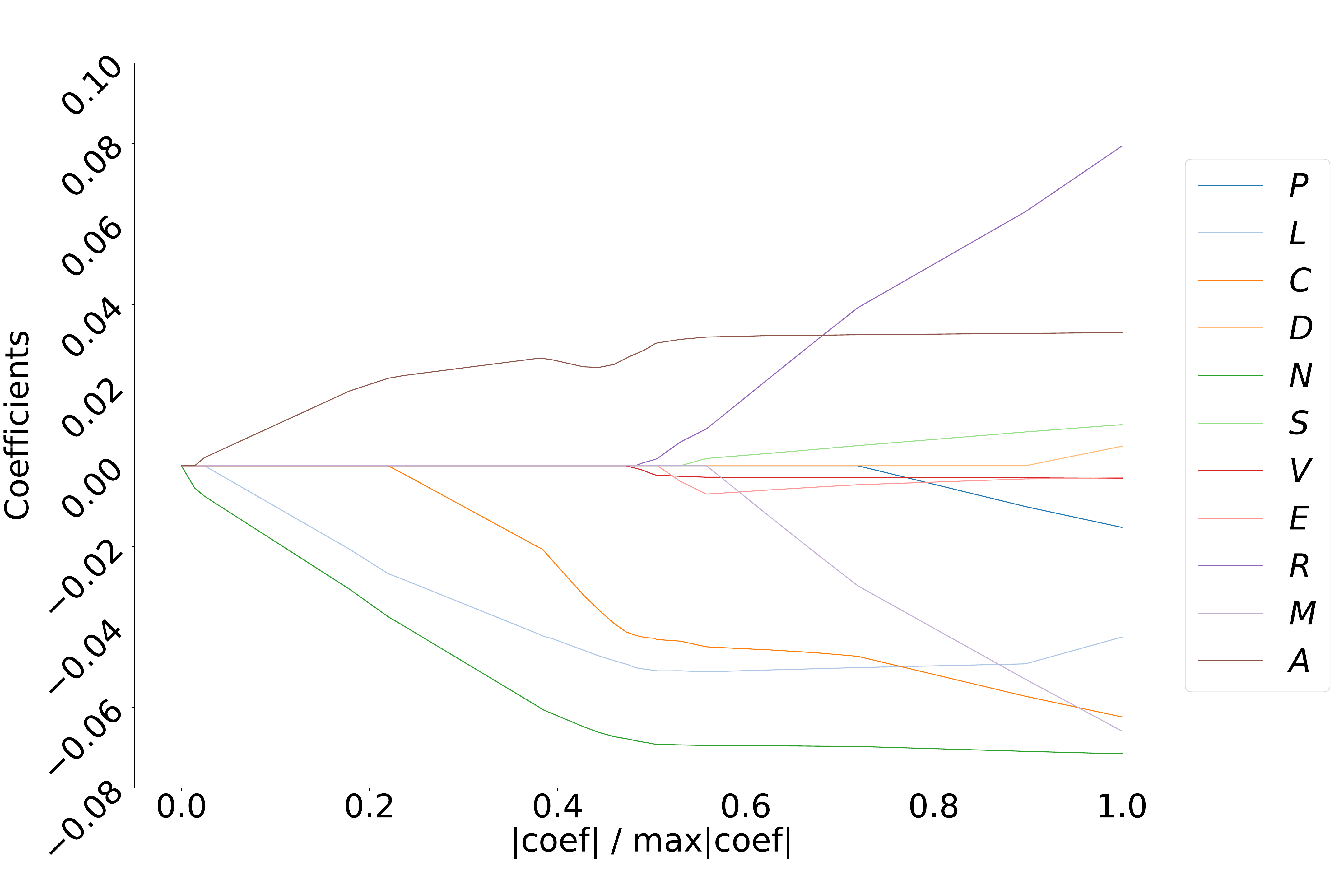}
  \caption{InstructBLIP (MSCOCO)}
  \label{fig:lasso_instuctblip_s}
\end{subfigure}%
\begin{subfigure}{.5\textwidth}
  \centering
  \includegraphics[width=\textwidth]{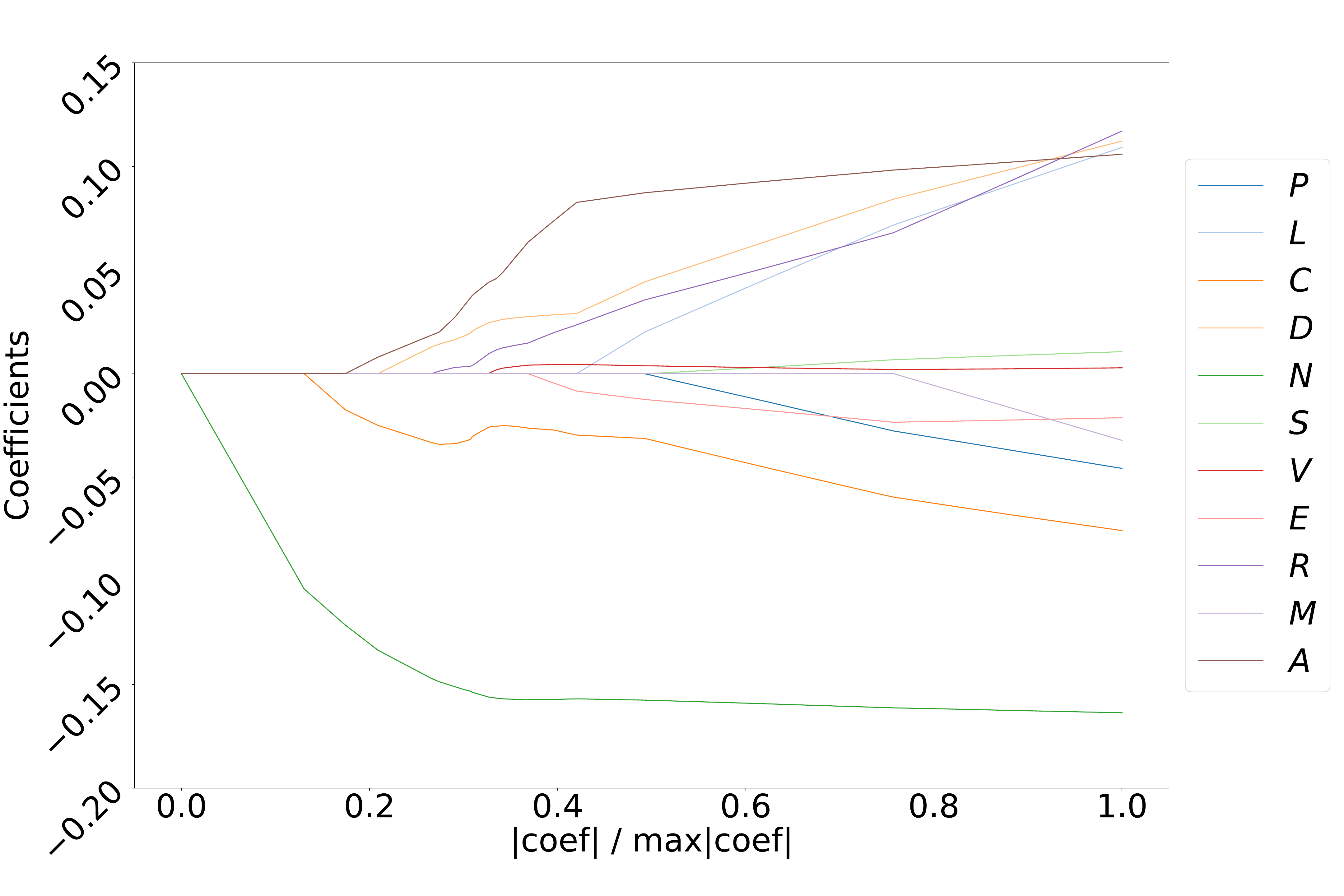}
  \caption{InstructBLIP (BDD100K)}
  \label{fig:lasso_instuct_blip_bdd}
\end{subfigure}
\begin{subfigure}{.5\textwidth}
  \centering
  \includegraphics[width=\textwidth]{fig/lasso_mPLUGOwl_sample.pdf}
  \caption{mPLUG-Owl (MSCOCO)}
  \label{fig:lasso_mplugowl_s}
\end{subfigure}
\begin{subfigure}{.5\textwidth}
  \centering
  \includegraphics[width=\textwidth]{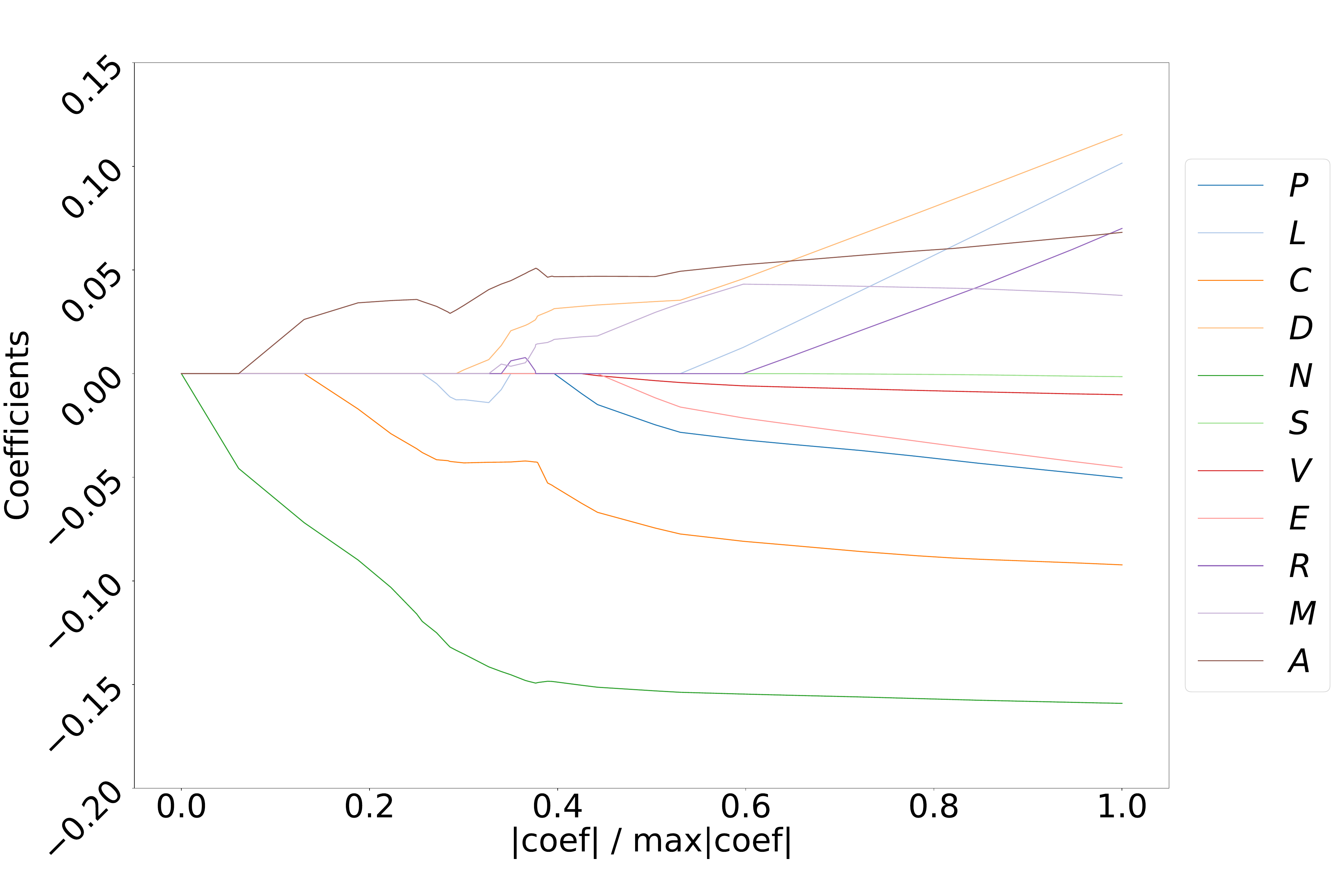}
  \caption{mPLUG-Owl (BDD100K)}
  \label{fig:lasso_mplug_owl_bdd}
\end{subfigure}
\begin{subfigure}{.5\textwidth}
  \centering
  \includegraphics[width=\textwidth]{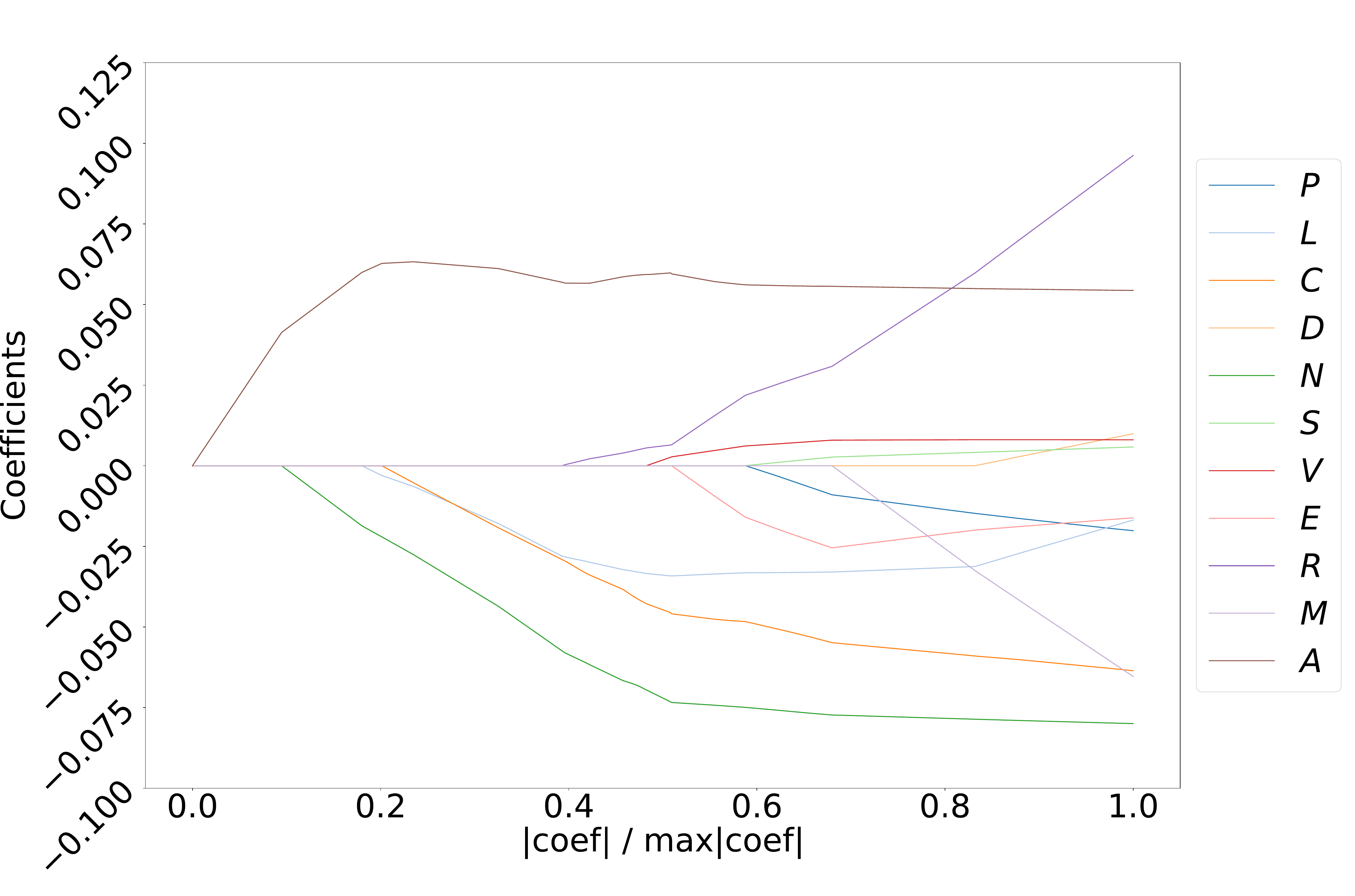}
  \caption{MiniGPT-4 (MSCOCO)}
  \label{fig:lasso_minigpt4_s}
\end{subfigure}
\begin{subfigure}{.5\textwidth}
  \centering
  \includegraphics[width=\textwidth]{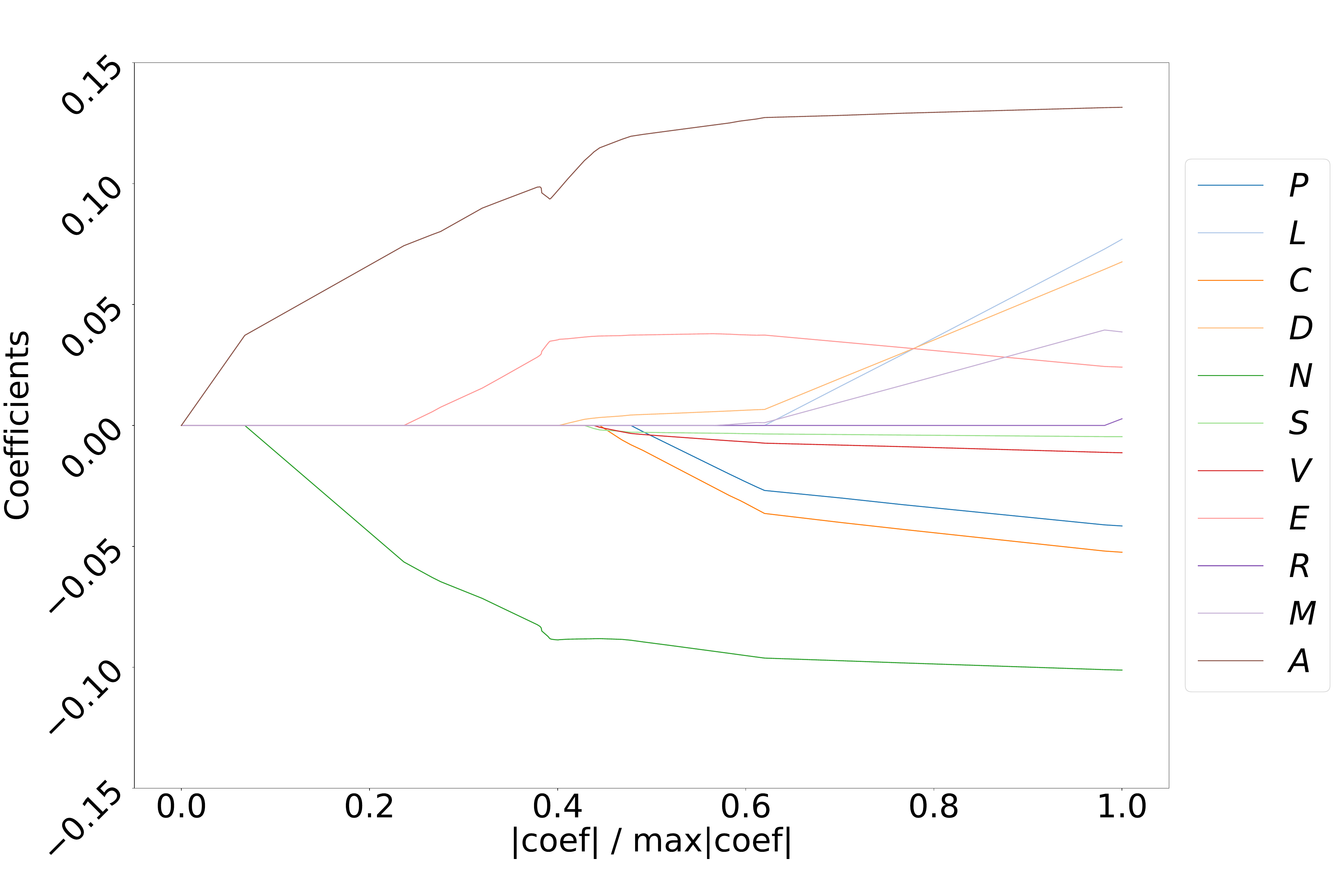}
  \caption{MiniGPT-4 (BDD100K)}
  \label{fig:lasso_minigpt4_bdd}
\end{subfigure}
\begin{subfigure}{.5\textwidth}
  \centering
  \includegraphics[width=\textwidth]{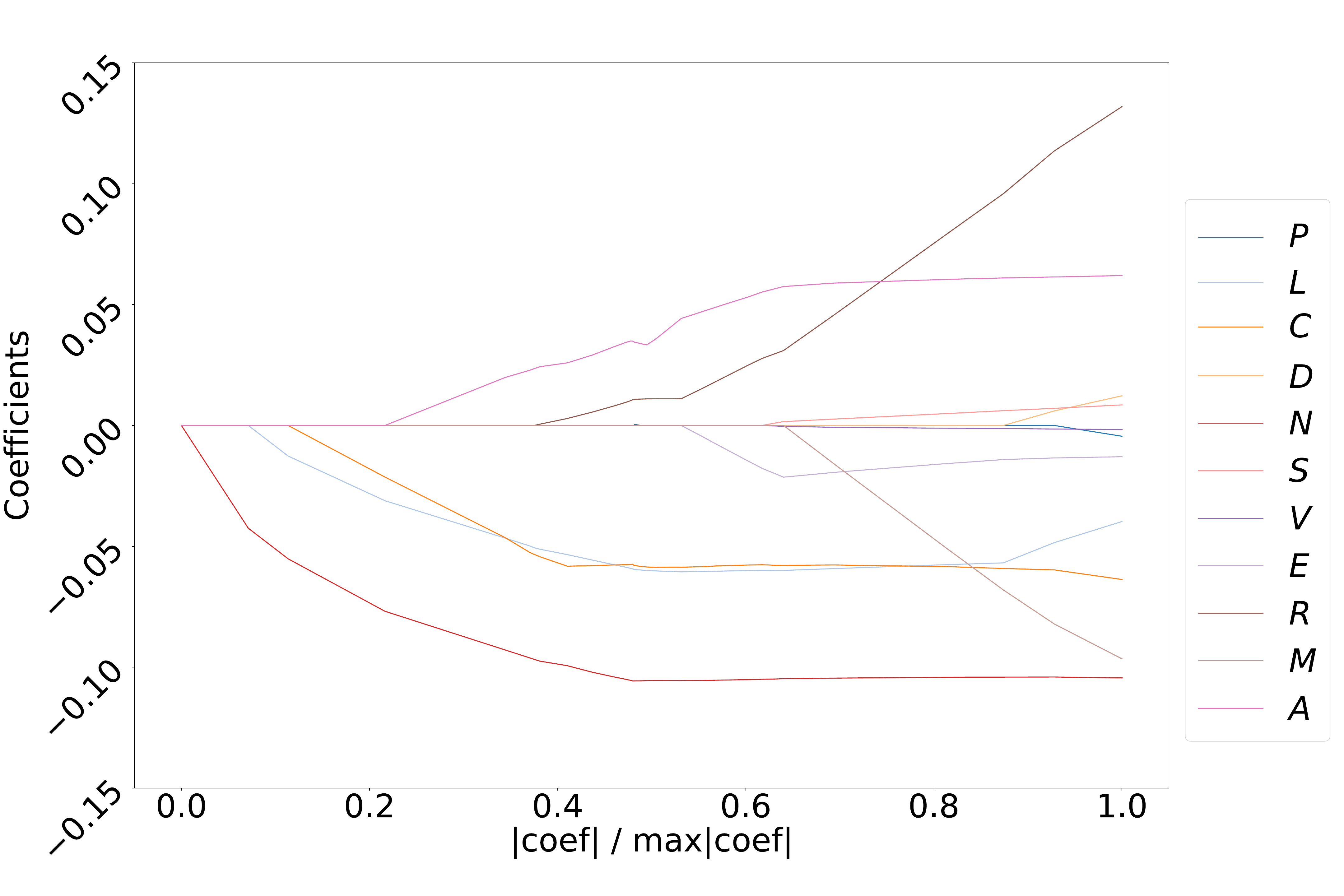}
  \caption{LLaVa (MSCOCO)}
  \label{fig:lasso_llava_s}
\end{subfigure}
\begin{subfigure}{.5\textwidth}
  \centering
  \includegraphics[width=\textwidth]{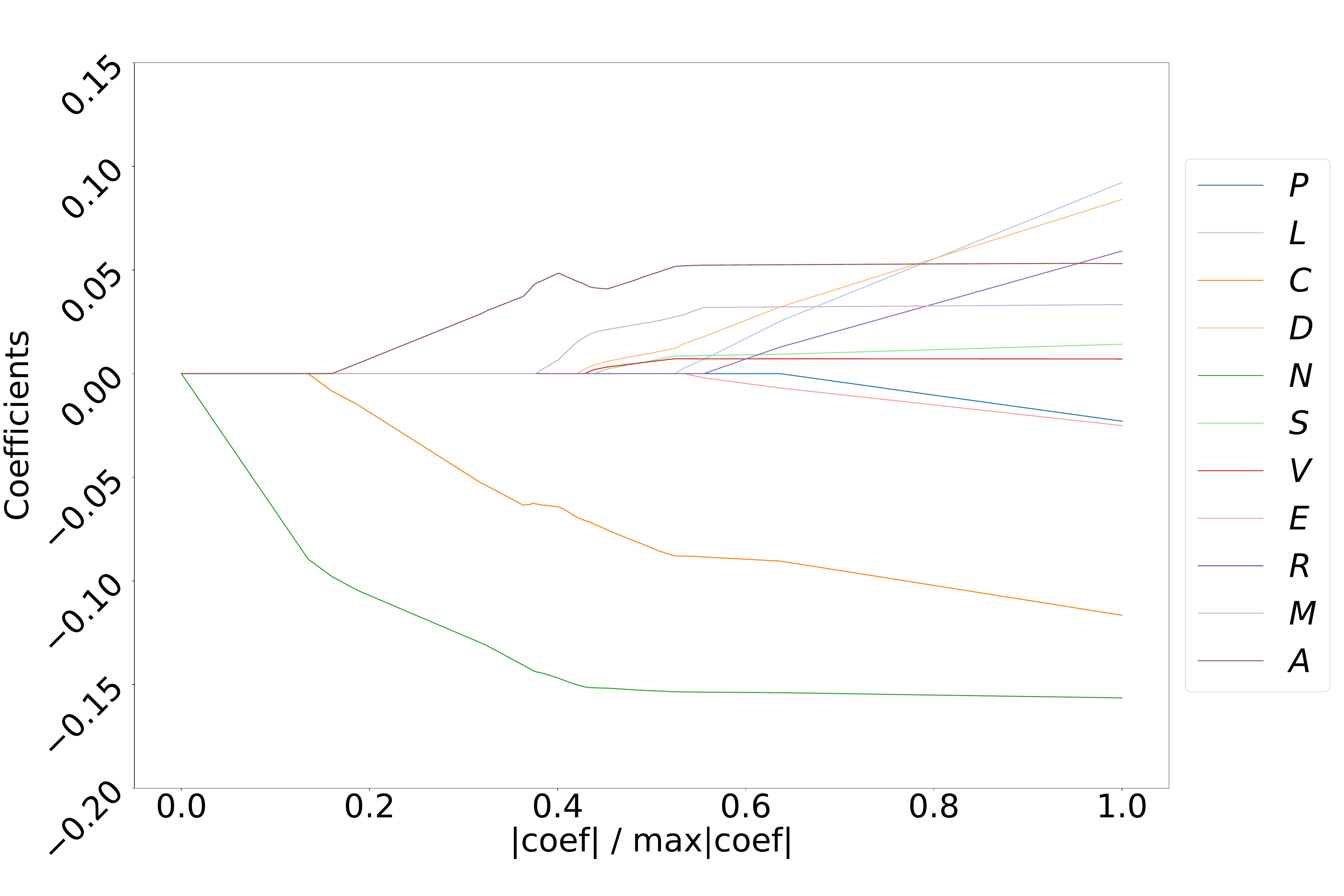}
  \caption{LLaVa (BDD100K)}
  \label{fig:lasso_llava_bdd}
\end{subfigure}
\caption{\textbf{LASSO Path based on Nucleus Sampling \cite{Holtzman.2020}.} LASSO path for $\mathcal{M}$ (\cref{eq:M}). $A$ denotes the maximum of the absolute values of all $G$ weight coefficients for the attention features $A^{g}, g=0,\dots,G-1$.}
\label{fig:lasso_full}
\end{figure*}

\begin{figure*}[t!]
\begin{subfigure}{.5\textwidth}
  \centering
  \includegraphics[width=\textwidth]{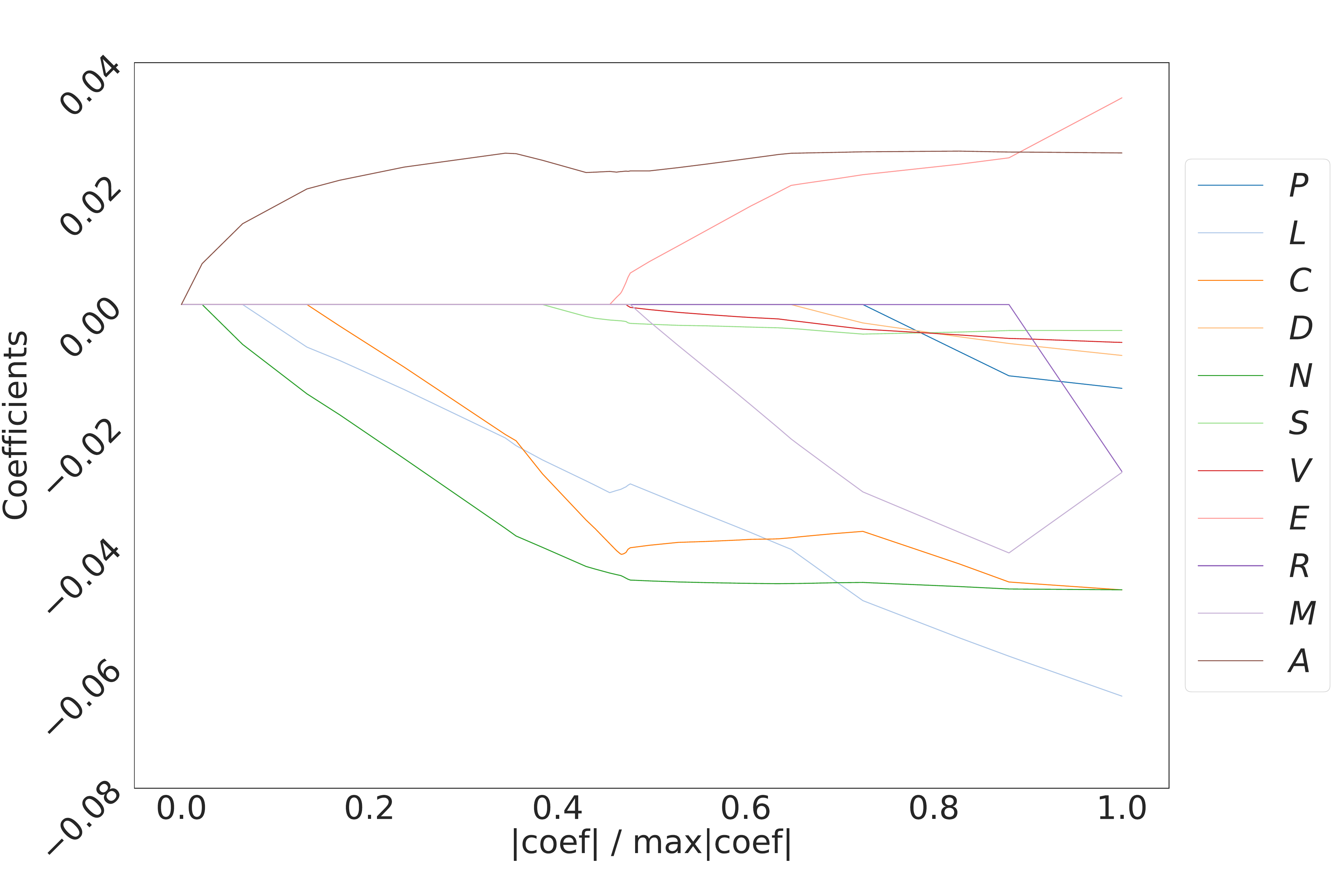}
  \caption{InstructBLIP (MSCOCO)}
  \label{fig:lasso_instuctblip_b}
\end{subfigure}%
\begin{subfigure}{.5\textwidth}
  \centering
  \includegraphics[width=\textwidth]{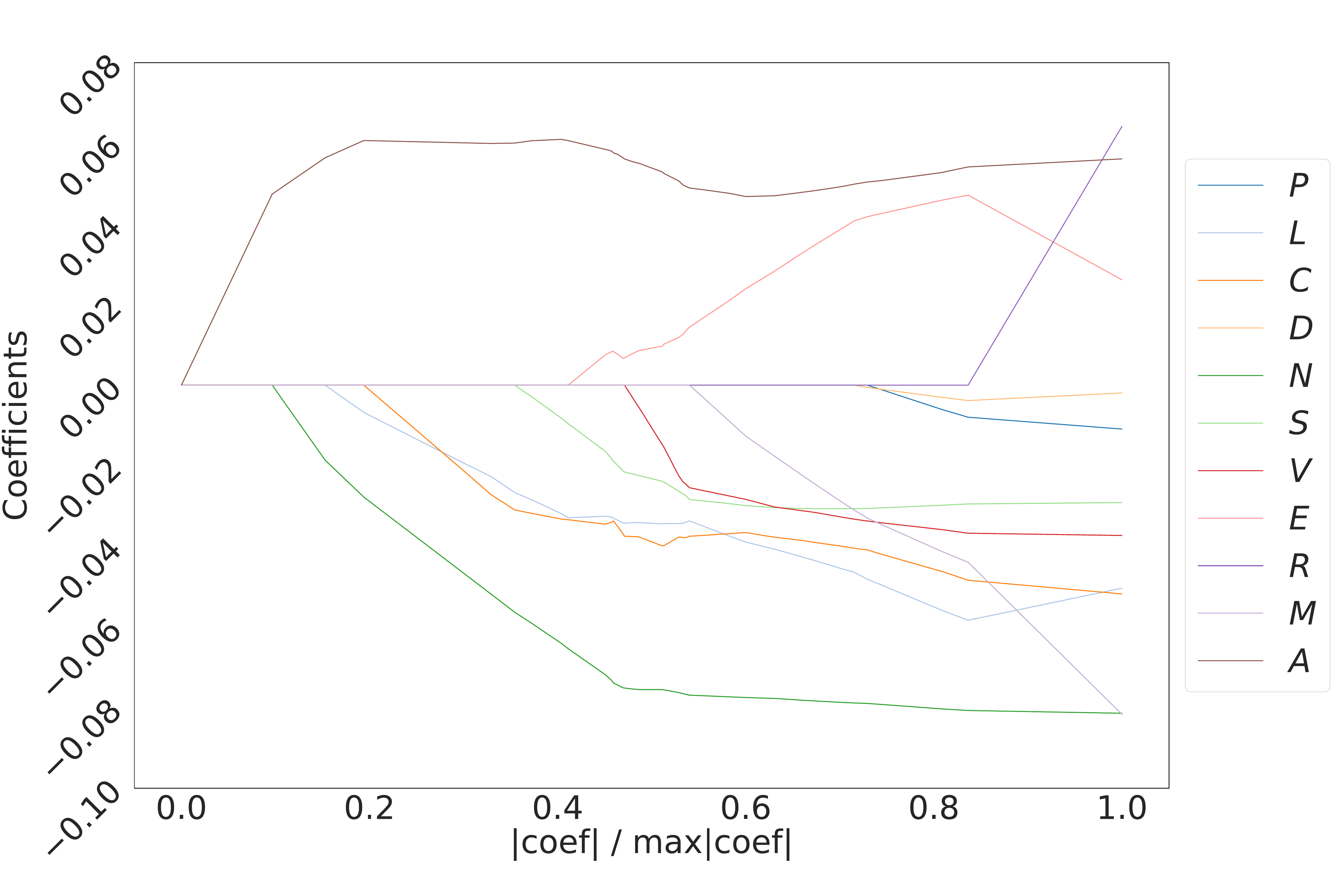}
  \caption{mPLUG-Owl (MSCOCO)}
  \label{fig:lasso_mplugowl_b}
\end{subfigure}
\caption{\textbf{LASSO Path based on Beam Search \cite{Bisiani.1987}.} LASSO path for $\mathcal{M}$ (\cref{eq:M}). $A$ denotes the maximum of the absolute values of all $G$ weight coefficients for the attention features $A^{g}, g=0,\dots,G-1$.}
\label{fig:lasso_beam}
\end{figure*}

We analyze the informative content included in the features proposed in \cref{subsec:metrics} for the considered LVLMs InstructBLIP, mPLUG-Owl, MiniGPT-4, and LLaVa 1.5 based on nucleus sampling \cite{Holtzman.2020}. For InstructBLIP and mPLUG-Owl we further investigated the considered features based on beam search \cite{Bisiani.1987}. To this end, we make use of the least absolute shrinkage and selection operator (LASSO) algorithm \cite{Efron.2004,Tibshirani.2018} to analyze the predictive power of the features considered. \cref{fig:lasso_full} and \cref{fig:lasso_beam} show the LASSO paths for the considered LVLMs and generation configurations. While the features absolute occurrence $N$ (\cref{eq:absocc}), mean absolute attention $A$ (\cref{eq:att}), log probability $L$ (\cref{eq:logp}) and cumulated log probability $C$ (\cref{eq:cumlogp}) are selected first in almost all settings, we observe different feature ranks between the sampling-based captions and beam search-based captions. While the coefficient for the entropy $E$ (\cref{eq:entropy}) is vanishing for the sampled captions (see \cref{fig:lasso_full}), it is selected as the sixth feature for the beam search-based captions (see \cref{fig:lasso_beam}).

\begin{figure*}[t!]
\begin{subfigure}{.5\textwidth}
  \centering
  \includegraphics[width=\textwidth]{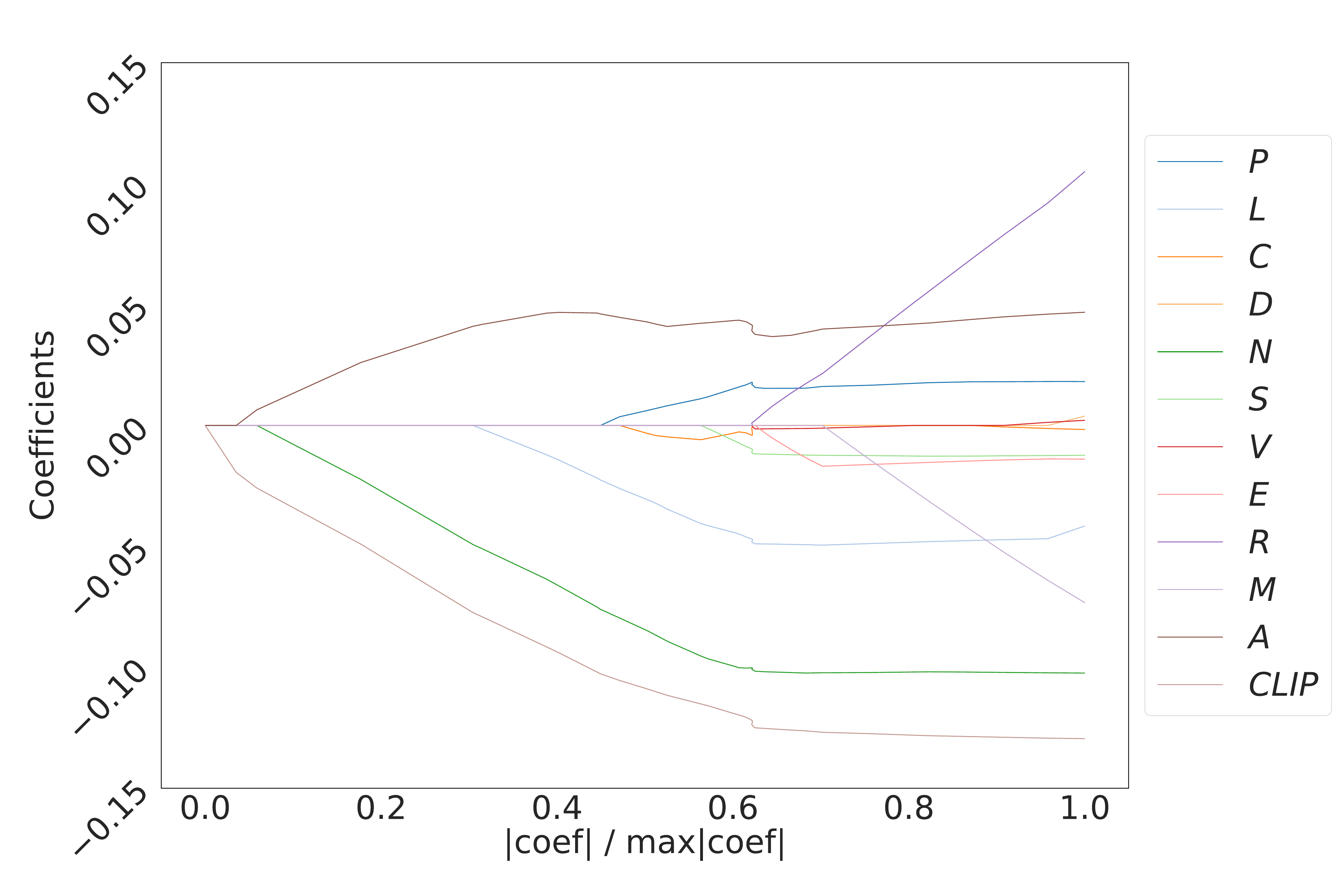}
  \caption{mPLUG-Owl (MSCOCO)}
  \label{fig:lasso_mplugowl_s_clip}
\end{subfigure}%
\begin{subfigure}{.5\textwidth}
  \centering
  \includegraphics[width=\textwidth]{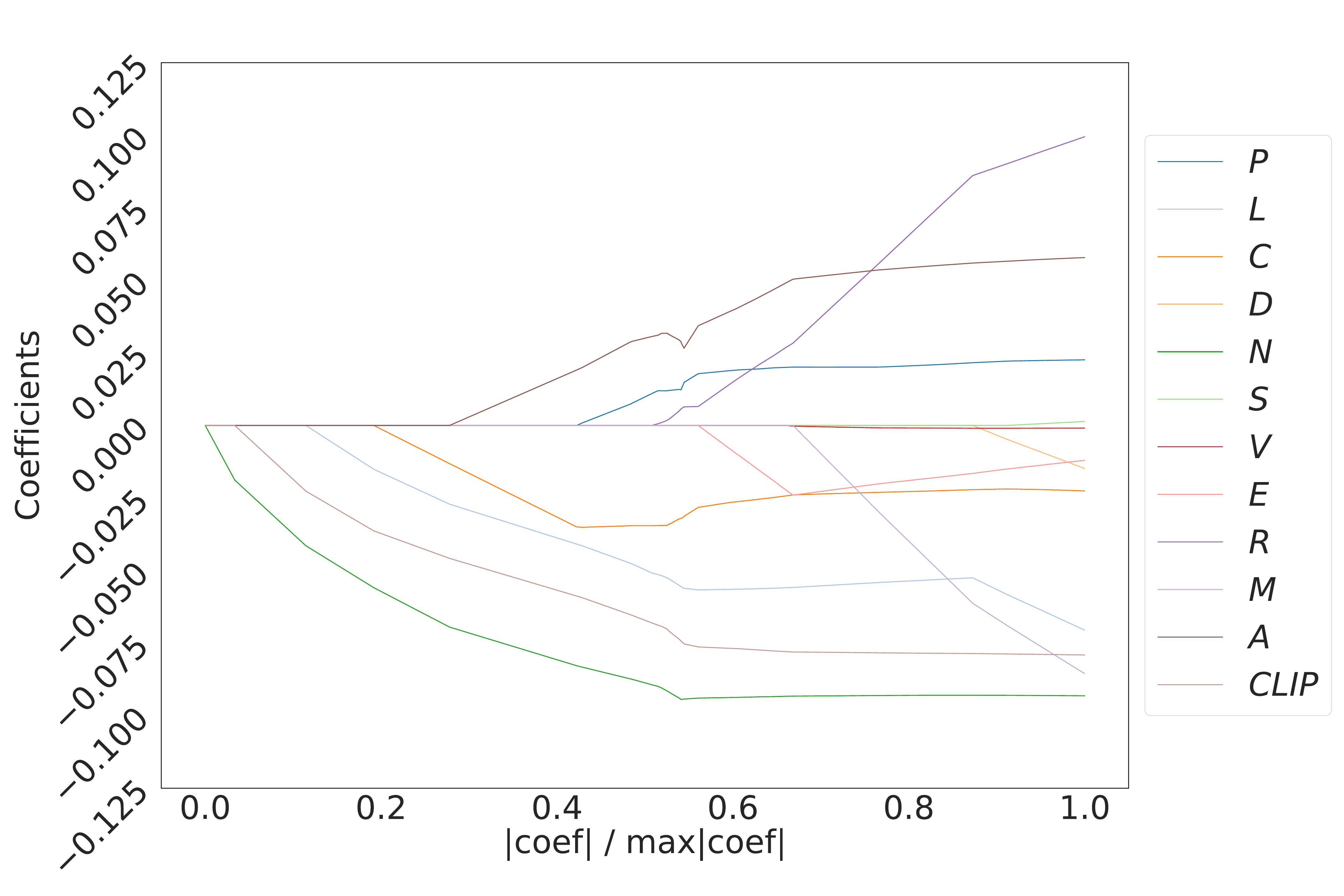}
  \caption{LLaVa (MSCOCO)}
  \label{fig:lasso_llava_clip}
\end{subfigure}
\caption{\textbf{LASSO Path based on the extended feature set $\mathcal{M}^{\mathrm{clip}}_{o_j}$.} $A$ denotes the maximum of the absolute values of all $G$ weight coefficients for the attention features $A^{g}, g=0,\dots,G-1$.}
\label{fig:lasso_full_clip}
\end{figure*}

\begin{figure*}[t!]
  \centering
  \includegraphics[width=\textwidth]{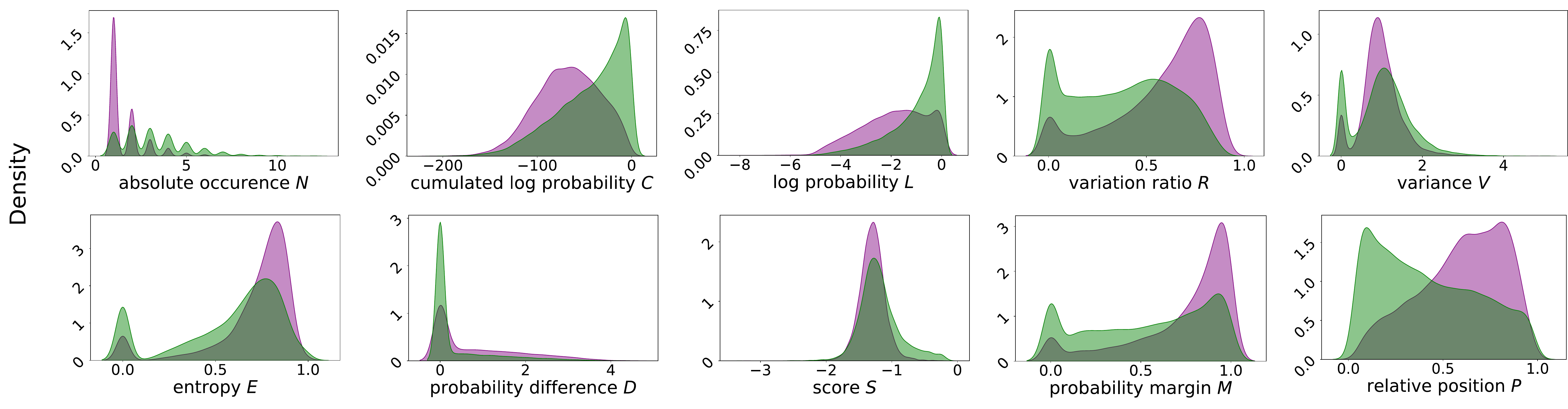}
  \caption{\textbf{Features.} Visualization of a selection of our proposed input features defined in \cref{subsec:metrics} for \colorbox{ao(english)!50}{true} and \colorbox{Mulberry!50}{hallucinated} objects.}
  \label{fig:metrics_full}
\end{figure*}

Moreover, we investigate the LASSO paths for the extended feature set $\mathcal{M}^{\mathrm{clip}}$ (\cref{eq:M_clip}) for the LVLMs sharing the vision encoder with CLIP \cite{Radford.2021}, that is, mPLUG-Owl and LLaVa in \cref{fig:lasso_full_clip}. For mPLUG-Owl, the $CLIP$ feature (\cref{eq:clip}) is selected as the first feature as we can see in \cref{fig:lasso_mplugowl_s_clip} followed by the mean absolute attention $A$ (\cref{eq:att}) the absolute occurrence $N$ (\cref{eq:absocc}) and the log probability $L$ (\cref{eq:logp}). For LLaVa, on the other hand, our proposed feature absolute occurrence $N$ (\cref{eq:absocc}) is selected first, underlining the informative content induced by this feature, followed by the $CLIP$ feature (\cref{eq:clip}), the log probability $L$ (\cref{eq:logp}) and cumulated log probability $C$ (\cref{eq:cumlogp}).

\subsection{Feature Visualization} \label{sec:app_metricvis}
\begin{figure*}[t!]
\begin{subfigure}{1.0\textwidth}
  \centering
  \includegraphics[width=\textwidth]{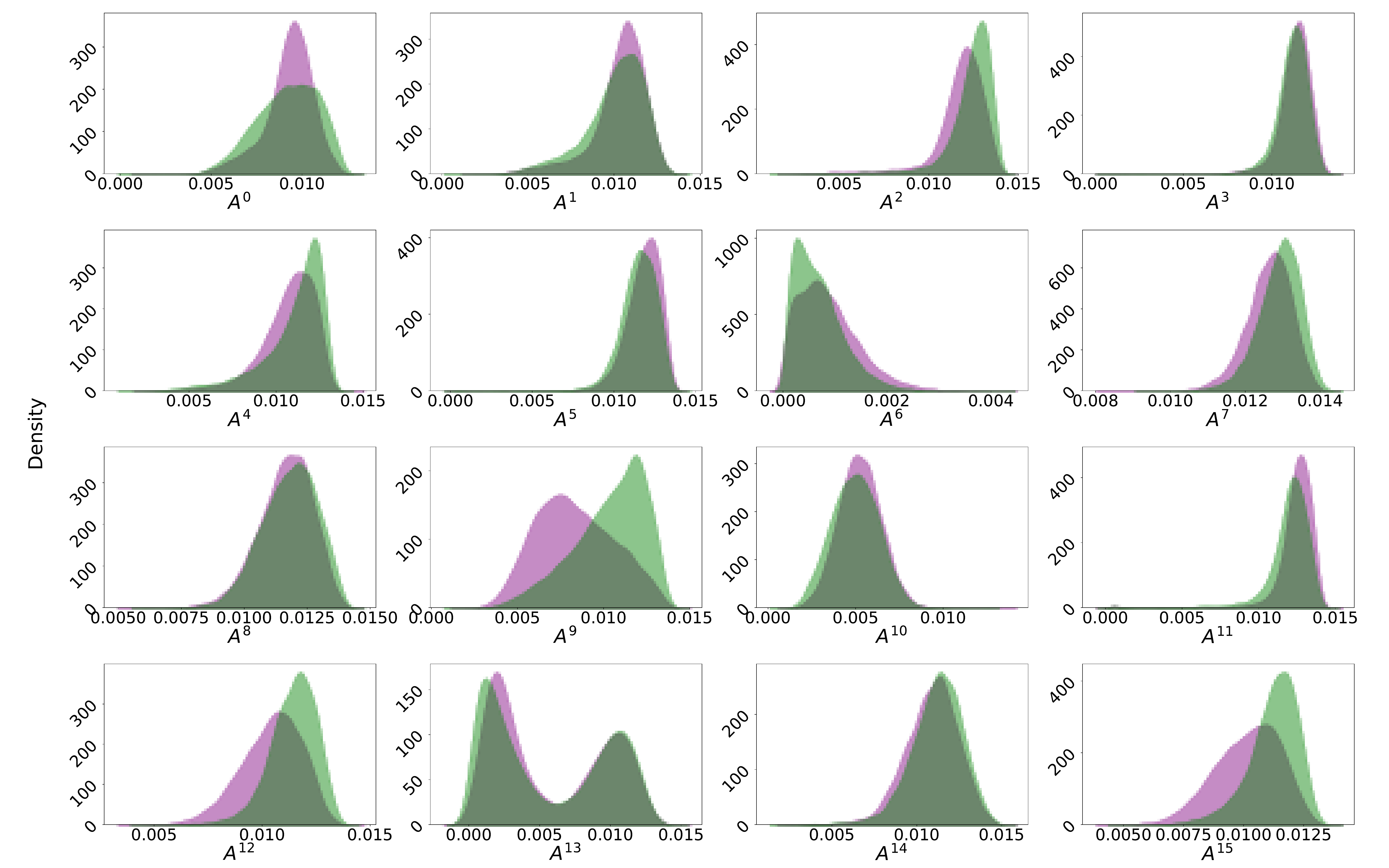}
\end{subfigure} \\
\begin{subfigure}{1.0\textwidth}
  \centering
  \includegraphics[width=\textwidth]{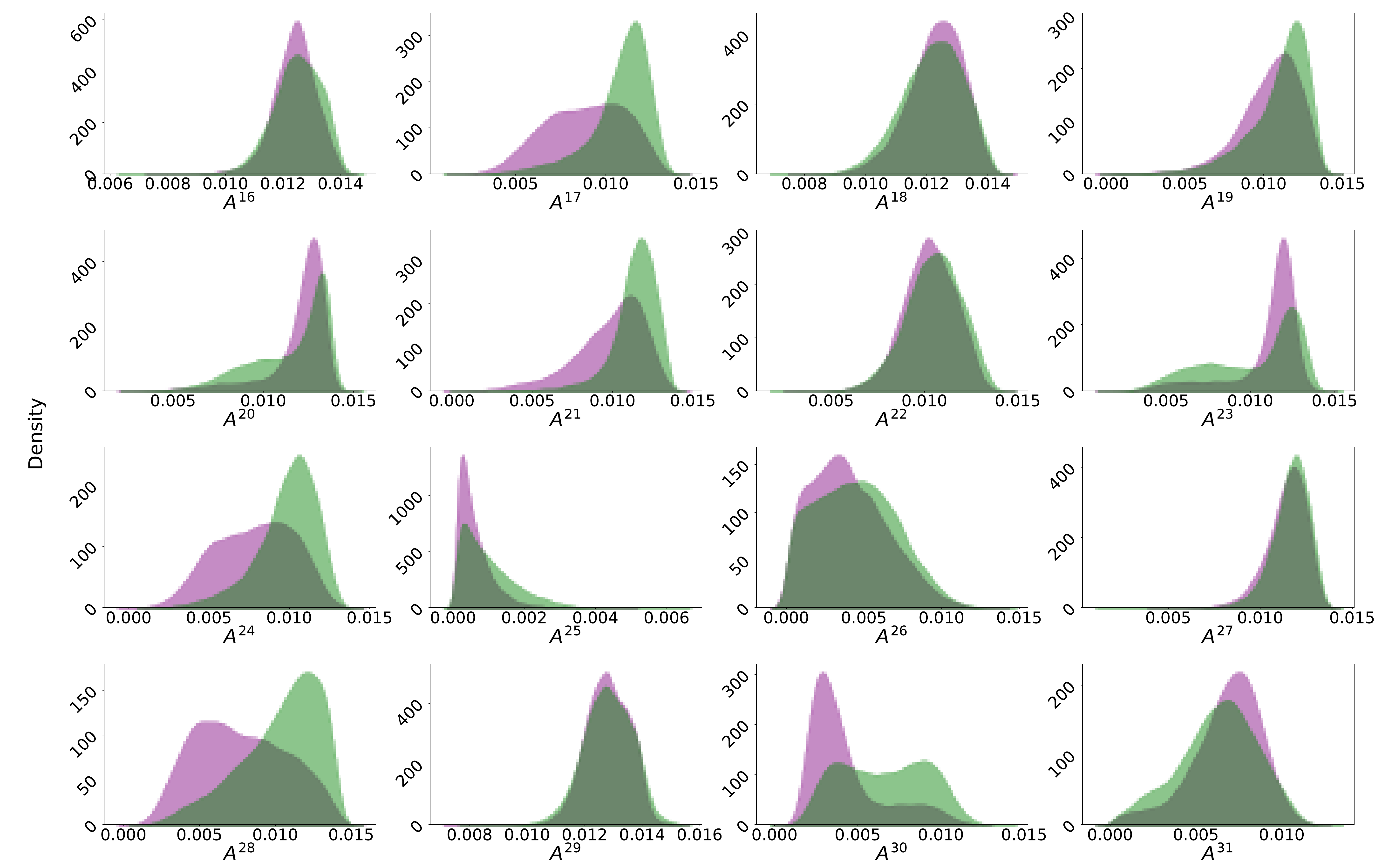}
\end{subfigure}%
  \caption{\textbf{Attention Features.} Visualization of the attention features \cref{eq:att} defined in \cref{subsec:metrics} for \colorbox{ao(english)!50}{true} and \colorbox{Mulberry!50}{hallucinated} objects.}
  \label{fig:att_full}
\end{figure*}

In order to provide a visual analysis of our proposed features (see \cref{subsec:metrics}), we plot our features for true and hallucinated objects in \cref{fig:metrics_full,fig:att_full}. While some of the attention features $A^g, g=0, ..., 31$ fully overlap for true and hallucinated objects (see \cref{fig:att_full}), for specific attention heads like $g=9,24$ and $28$ the respective features distinguish between hallucinated and true objects. We can also see in \cref{fig:metrics_full} that our proposed feature absolute occurrence $N$ (\cref{eq:absocc}) has a high density for $N=1$ for hallucinated objects, that is, if the object is only mentioned once in the image caption. True objects usually occur several times in the image caption. However, we want to emphasize the importance of our statistical analysis based on the LASSO algorithm in \cref{sec:app_lasso}. While in \cref{fig:metrics_full}, the relative position $P$ (\cref{eq:relpos}) might look like a better feature to classify between hallucinated and true objects than the log probability $L$ (\cref{eq:logp}) or the cumulated log probability $C$ (\cref{eq:cumlogp}), our analysis in the previous section shows that the relative position $P$ only adds minor information to the classifier reflected by a rather small coefficient compared to the log probability $L$ and cumulated log probability $C$ (see \cref{fig:lasso_full}).

\end{document}